\documentclass[twoside,11pt]{article}

\usepackage{algorithm}
\usepackage{algorithmic}

\usepackage{amsmath}
\usepackage{float}
\RequirePackage{bm}
\usepackage{multirow}
\usepackage{amsmath}
\usepackage{wrapfig}
\usepackage{dsfont}
\usepackage{tabularx}
\usepackage{subfigure}
\usepackage{enumitem}
\usepackage{color}

\usepackage{jmlr2e}
\usepackage{mkolar_definitions}

\usepackage{mathtools}
\mathtoolsset{showonlyrefs=true}

\let\hat\widehat

\graphicspath{{fig/}}

\usepackage{lastpage}
\jmlrheading{21}{2020}{1-\pageref{LastPage}}{6/19; Revised
2/20}{3/20}{19-496}{Ming Yu, Varun Gupta, and Mladen Kolar}

\ShortHeadings{Low-rank Topic-Based Model for Information Cascades}{Yu, Gupta, and Kolar}
\firstpageno{1}

\begin{document}

\title{Estimation of a Low-rank Topic-Based Model for Information Cascades}

\author{\name Ming Yu \email mingyu@chicagobooth.edu \\
       \name Varun Gupta \email varun.gupta@chicagobooth.edu  \\
       \name Mladen Kolar \email mladen.kolar@chicagobooth.edu  \\
       \addr Booth School of Business \\
       The University of Chicago\\
       Chicago, IL 60637, USA
}

\editor{Boaz Nadler}

\maketitle

\begin{abstract}%
  We consider the problem of estimating the latent structure of a
  social network based on the observed information diffusion
  events, or {\it cascades}, where the observations for a given cascade 
  consist of only the timestamps of infection for infected nodes but 
  not the source of the infection.
  Most of the existing work on this problem has focused on estimating a
  diffusion matrix without any structural assumptions on it.  In this
  paper, we propose a novel model based on the intuition that an
  information is more likely to propagate among two nodes if they are
  interested in similar topics which are also prominent in the information content.
  In particular, our model endows each node with an influence vector
  (which measures how authoritative the node is on each topic) and a
  receptivity vector (which measures how susceptible the node is for each
  topic). We show how this node-topic structure can be estimated from
  the observed cascades, and prove the consistency of the estimator. 
  Experiments on synthetic and real
  data demonstrate the improved performance and better
  interpretability of our model compared to existing state-of-the-art
  methods.
\end{abstract}



\section{Introduction}

The spread of information in online web or social networks, the
propagation of diseases among people, as well as the diffusion of
culture among countries are all examples of information diffusion
processes or cascades.  In many of the applications, it is common to
observe the spread of a cascade, but not the underlying network
structure that facilitates the spread. For example, marketing data sets capture the times of purchase  of products by consumers, but not whether the consumer was influenced by a recommendation of a friend or an advertisement on TV; we
can observe when a person falls ill, but we cannot observe who
infected him/her.  In all these settings, we can observe the
propagation of information but cannot observe the way they propagate.

There is a vast literature on recovering the underlying network
structure based on the observations of information diffusion.  A
network is represented by a diffusion matrix that characterizes
connections between nodes, that is, the diffusion matrix gives
weight/strength to the arcs between all ordered pairs of vertices.
\citet{Gomez-Rodriguez2011Uncovering} propose a continuous time diffusion
model and formulate the problem of recovering the underlying network
diffusion matrix by maximizing the log-likelihood function.  The model
of \citet{Gomez-Rodriguez2011Uncovering} imposes no structure among nodes
and allows for arbitrary diffusion matrices. As a modification of this
basic model, \citet{du2013uncover} consider a more sophisticated
topic-sensitive model where each information cascade is associated
with a topic distribution on several different topics.  Each topic is
associated with a distinct diffusion matrix and the diffusion matrix
for a specific cascade is a weighted sum of these diffusion matrices
with the weights given by the topic distribution of the cascade.  This
model can capture our intuition that news on certain topics (for example,
information technology) may spread much faster and broader than some
others (for example, military).  However, since the diffusion matrix for each
topic can be arbitrary, the model fails to capture the intuition that
nodes have intrinsic topics of interest.

In this paper, we propose a novel mathematical model that incorporates
the node-specific topics of interest.  Throughout the paper we use the
diffusion of news among people as an example of cascades for
illustrative purposes.  An item of news is usually focused on one or a
few topics (for example, entertainment, foreign policy, health), and is more
likely to propagate between two people if both of them are interested
in these same topics.  Furthermore, a news item is more likely to be
shared from node 1 to node 2 if node 1 is influential/authoritative in
the topic, and node 2 is receptive/susceptible to the topic.  Our
proposed mathematical model is able to capture this intuition.  We
show how this node-topic structure (influence and receptivity) can be
estimated based on observed cascades with a theoretical guarantee.
Finally, on the flip side, after obtaining such a network structure,
we can use this structure to assign a topic distribution to a new
cascade.  For example, an unknown disease can be classified by looking
at its propagation behavior.

To the best of our knowledge, this is the first paper to leverage
users' interests for recovering the underlying network structure from
observed information cascades.  Theoretically, we prove that our
proposed algorithm converges linearly to the true model parameters up
to statistical error; experimentally, we demonstrate the scalability
of our model to large networks, robustness to overfitting, and better
performance compared to existing state-of-the-art methods on both
synthetic and real data.  While existing algorithms output a large
graph representing the underlying network structure, our algorithm
outputs the topic interest of each node, which provides better
interpretability.  This structure can then be used to predict future
diffusions, or for customer segmentation based on interests.  It can
also be applied to build recommendation systems, and for marketing applications such as targeted advertising, which is impossible for existing works.

A conference version of this paper was presented in the
2017 IEEE International Conference on Data Mining (ICDM) series
\citep{yu2017influence}. Compared to the conference version, in this
paper we extend the results in the following ways:
(1) we introduce a new penalization method and a new algorithm in Section \ref{sec:Optimization};
(2) we build theoretical result for our proposed algorithm in Section \ref{sec:theoretical};
(3) we discuss several variants and applications of our model in Section \ref{sec:Variants};
(4) we evaluate the performance of our algorithm on a new data set in Section \ref{sec:Memetracker}.

\subsection{Related Work}

A large body of literature exists on recovery of latent network
structure based on observed information diffusion cascades
\citep{kempe2003maximizing, gruhl2004information}.  See
\citet{guille2013information} for a survey.
\citet{Pouget-Abadie2015Inferring} introduce a Generalized Linear Cascade
Model for discrete time. Alternative approaches to analysis
of discrete time networks have been considered in
\citep{eagle2009inferring,Song2009KELLER,song09time,kolar10nonparametric,kolar2010estimating,kolar2011time,kolar10estimating,Lozano2010Block,netrapalli2012learning,Wang2014Inference,gao2016periodic,Lu2015Posta}.

In this paper we focus on network inference under the continuous-time
diffusion model introduced in \cite{Gomez-Rodriguez2011Uncovering},
where the authors formulate the network recovery problem as a convex
program and propose an efficient algorithm ($\textbf{NetRate}$) to
recover the diffusion matrix.  In a follow-up work,
\cite{Gomez-Rodriguez2010Inferring} look at the problem of finding the
best $K$ edge graph of the network. They show that this problem is
NP-hard and develop \textbf{NetInf} algorithm that can find a
near-optimal set of $K$ directed edges.
\cite{Gomez-Rodriguez2013Structure} consider a dynamic network
inference problem, where it is assumed that there is an unobserved
dynamic network that changes over time and propose \textbf{InfoPath}
algorithm to recover the dynamic network. \cite{du2012learning} relax
the restriction that the transmission function should have a specific
form, and propose \textbf{KernelCascade} algorithm that can infer the
transmission function automatically from the data.
Specifically, to better capture the heterogeneous influence among nodes, each pair of nodes can have a different type of transmission model.
\cite{zhou2013learning} use multi-dimensional Hawkes processes to
capture the temporal patterns of nodes behaviors. By optimizing
the nuclear and $\ell_1$ norm simultaneously, \textbf{ADM4} algorithm
recovers the network structure that is both low-rank and sparse.
\cite{myers2012information} consider external influence in the model:
information can reach a node via the links of the social network or
through the influence of external sources.  \cite{myers2012clash}
further assume interaction among cascades: competing cascades
decrease each other's probability of spreading, while cooperating
cascades help each other in being adopted throughout the network.
\cite{Gomez-Rodriguez2016Estimating} prove a lower bound on the
number of cascades needed to recover the whole network structure
correctly.  \cite{he2015hawkestopic} combine Hawkes processes and
topic modeling to simultaneously reason about the information
diffusion pathways and the topics of the observed text-based cascades.
Other related works include \citep{bonchi2011influence, liu2012time,
  du2013scalable, Gomez-Rodriguez2012Influence, jiang2014evolutionary,
  zhang2016dynamics}.

The work most closely related to ours is \cite{du2013uncover}, where
the authors propose a topic-sensitive model that modifies the basic
model of \cite{Gomez-Rodriguez2011Uncovering} to allow cascades with
different topics to have different diffusion rates.
However, this topic-sensitive model still fails to account for the
interaction between nodes and topics.

\subsection{Organization of the Paper}

In Section~\ref{sec:Background} we briefly review the basic
continuous-time diffusion network model introduced in
\citet{Gomez-Rodriguez2011Uncovering} and the topic-sensitive model
introduced in \citet{du2013uncover}.  We propose our
influence-receptivity model in Section~\ref{sec:NodeTopic}.
Section~\ref{sec:Optimization} details two optimization algorithms.
Section~\ref{sec:theoretical} provides theoretical results for the
proposed algorithm.  In Section~\ref{sec:Variants} we discuss
extensions of our model. Sections~\ref{sec:ExperimentsSynthetic} and
\ref{sec:ExperimentsReal} present experimental results on synthetic
data set and two real world data sets, respectively.  We conclude in
Section~\ref{sec:conclusion}.

\subsection{Notation}

We use $p$ to denote the number of nodes in a network and $K$ to
denote the number of topics. The number of observed cascades is
denoted as $n$.  We use subscripts $i, j \in\{1, \ldots, p\}$ to index
nodes; $k \in \{1,\ldots, K\}$ to index topics; and $c$ to index each
cascade.  For any matrix $A$, we use $\|A\|_2$ and $\|A\|_F$ to denote
the matrix spectral norm and Frobenius norm, respectively.  Moreover,
$\|A\|_0 = \big|(i,j): A_{ij} \neq 0\big|$ denotes the number of
nonzero components of a matrix.  The operation $[A]_+$ keeps only
nonnegative values of $A$ and puts zero in place of negative values.
For a nonnegative matrix $A$, the operation $\text{Hard}(A, s)$ keeps
only the $s$ largest components of $A$ and zeros out the rest of the
entries.  We use $S = \text{supp}(A) = \cbr{(i,j): A_{ij} \neq 0}$ to
denote the support set of matrix $A$ (with an analogous definition for
a vector).  For any matrix $A$ and support set $S$, we denote $[A]_S$ as
the matrix that takes the same value as $A$ on $S$, and zero
elsewhere.  For any matrices $A$ and $B$, denote
$\langle A, B \rangle = \tr(A^\top B)$ as the matrix inner product and
$\langle A, B \rangle_S = \tr \big([A]_S^\top \cdot [B]_S \big)$ as
the inner product on the support $S$ only.


\section{Background}
\label{sec:Background}

We briefly review the basic continuous time diffusion network model
introduced in \cite{Gomez-Rodriguez2011Uncovering} in
Section~\ref{sec:Basicmodel}.  The topic-sensitive model introduced as
a modification of the basic model in \cite{du2013uncover}
is reviewed in Section~\ref{sec:topicsensitive}.

\subsection{Basic Cascade Model}
\label{sec:Basicmodel}

\begin{figure}[htbp]
\begin{center}
\includegraphics[width=0.35\textwidth]{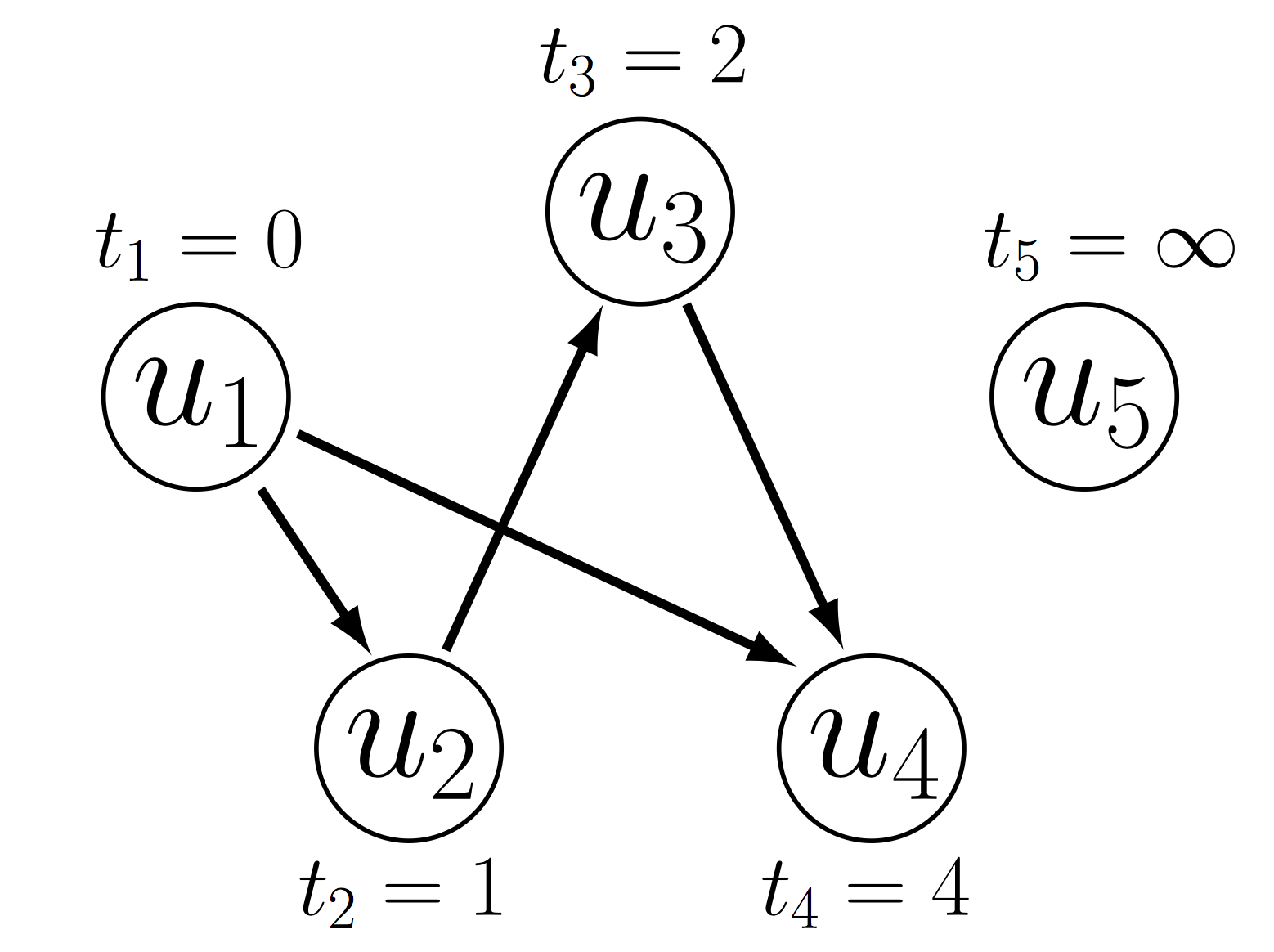}
\caption{An illustrative example of one cascade.}
\label{fig:cascade}
\end{center}
\end{figure}

\paragraph{\emph{Example.}}We first provide an illustrative example of a cascade in Figure \ref{fig:cascade}.
Here we have 5 nodes in the network, termed $u_1$ to $u_5$.
At time $t_1 = 0$, node $u_1$ knows some information, and starts the information diffusion process.
Node $u_2$ gets ``infected" at time $t_2 = 1$.
The process continues, and node $u_3$, $u_4$ become aware of the information
at times $t_3 = 2$ and $t_4 = 4$, respectively.
Node $u_5$ never gets infected, so we write $t_5 = \infty$.
The arrows in Figure \ref{fig:cascade} represent the underlying network.
However, we only observe the times at which each node gets infected: $t = [0, 1,2,4,\infty]$.

\paragraph{\emph{Network structure and cascade generating process.}}

The model of \cite{Gomez-Rodriguez2011Uncovering} assumes that the
underlying network is composed of $p$ nodes and uses a non-negative
diffusion matrix $A = \{\alpha_{ji}\}$ to parameterize the edges among
them.  The parameter $\alpha_{ji}$ measures the transmission rate from
$j$ to $i$, where a larger $\alpha_{ji}$ means stronger connection
from $j$ to $i$.  The absence of $j \to i$ edge is denoted by
$\alpha_{ji} = 0$. For every node $i$, self infection is not
considered and $\alpha_{ii} = 0$. A cascade based on the model and
network here is generated in the following way.  At the beginning, at time $0$, one of the $p$ nodes is infected as a source node.
When a node $j$ is infected, it samples a time at which it infects
other uninfected nodes it is connected to.  The transmission time
$\tau_{ji}$ from node $j$ to $i$ follows a random distribution with a
density $\ell(\tau;\alpha_{ji})$ for $\tau \geq 0$ (this density is
called the {\it transmission function/kernel}).  A node $i$ is
infected the first time one of the nodes which can reach $i$ infects
it. After being infected, node $i$ becomes a new source and begins to
infect other nodes by following the same procedure and sampling the
transmission times to other uninfected nodes that it can reach.
An infected node continues to infect additional nodes after infecting one of its neighbor nodes.

The model assumes an observation window of length $T$ time units since
the infection of the source node; nodes that are not infected until
time $T$ are regarded as uninfected.  We write
$\ell(t_i \mid t_j;\alpha_{ji}) = \ell(t_i-t_j;\alpha_{ji})$ to
indicate the density that $i$ is infected by $j$ at time $t_i$ given
that $j$ is infected at time $t_j$, parameterized by $\alpha_{ji}$.
The transmission times of each infection are assumed to be
independent, and a node remains infected in the whole process once it
is infected.

\paragraph{\emph{Data.}} In order to fit parameters of the model above, we
assume that there are $n$ independent cascades denoted by the set
$C^n = \{ t^1, \ldots, t^n\}$.  A cascade $c$ is represented by
$ t^c$, which is a $p$-dimensional vector
$ t^c = (t_1^c, \ldots, t_p^c)$ indicating the time of infection of
the $p$ nodes; $t_i^c \in [0, T^c] \bigcup \{\infty\}$ with $T^c$
being the observation window for the cascade $c$.  Although not
necessary, for notational simplicity we assume $T^c = T$ for all the
cascades.  For an infected node, only the first infected time is
recorded even if it is infected by multiple neighbors.  For the source
node $i$, $t_i^c = 0$, while node uninfected up to time $T$ we use the convention $t_i^c = \infty$.
Moreover, the network structure is assumed to be \emph{static} and not change while the $n$ different cascades are observed.

\paragraph{\emph{Likelihood function.}}

The likelihood function of an observed cascade $t$ is given by
\begin{equation}
\begin{aligned}
\ell(  t;  A) = \prod\limits_{t_i \leq T}  \prod\limits_{t_m > T} S(T\mid t_i;\alpha_{im})  \times  \bigg[ \prod\limits_{k:t_k < t_i}S(t_i\mid t_k;\alpha_{ki}) \sum_{j:t_j<t_i}H(t_i \mid t_j;\alpha_{ji}) \bigg] ,
\label{likeli}
\end{aligned}
\end{equation}
where
$S(t_i\mid t_j;\alpha_{ji}) = 1 - \int_{t_j}^{t_i} \ell(t-t_j;\alpha_{ji})\,dt$ is the survival function and
$H(t_i|t_j;\alpha_{ji}) = \ell(t_i-t_j;\alpha_{ji}) / S(t_i|t_j;\alpha_{ji})$ is the hazard function
\citep{Gomez-Rodriguez2011Uncovering}.
Note that the likelihood function consists of two probabilities. The
first one is the probability that an uninfected node ``survives'' given
its infected neighbors; the second one is the density that an infected
node is infected at the specific observed time.

The transmission function affects the behavior of a cascade. Some
commonly used transmission functions are exponential, Rayleigh, and
power-law distributions \citep{Gomez-Rodriguez2011Uncovering}.  For
exponential transmission, the diffusion rate reaches its maximum value
at the beginning and then decreases exponentially. Because of this
property, it can be used to model information diffusion on internet or
a social network, since (breaking) news usually spread among people
immediately, while with time a story gradually becomes unpopular.
The exponential transmission function is given by
\begin{equation}
\label{eq:def_exponential_transmission}
\ell(\tau;\alpha_{ji}) = \alpha_{ji} \cdot \exp(-\alpha_{ji}\tau)
\end{equation}
for
$\tau \geq 0$ and $\ell(\tau;\alpha_{ji}) = 0$ otherwise. We then have
$S(t+\tau\mid t;\alpha_{ji}) = \exp(-\alpha_{ji}\tau)$ and
$H(t+\tau\mid t;\alpha_{ji}) = \alpha_{ji}$.
As a different example, with the Rayleigh transmission function
the diffusion rate is small at the beginning; it then rises to a peak and then drops.
It can be used to model citation networks, since it usually takes some
time to publish a new paper and cite the previous paper. New papers
then gradually become known by researchers.
The Rayleigh transmission function is given as
$$
\ell(\tau;\alpha_{ji}) = \alpha_{ji} \tau \cdot \exp\Big(-\frac
12\alpha_{ji}\tau^2\Big)
$$
for $\tau \geq 0$ and $\ell(\tau;\alpha_{ji}) = 0$ otherwise. We then
have $S(t+\tau\mid t;\alpha_{ji}) = \exp(-\frac 12\alpha_{ji}\tau^2)$ and
$H(t+\tau\mid t;\alpha_{ji}) = \alpha_{ji}\tau$.
We will use these two transmission functions in
Section \ref{sec:ExperimentsReal}
for modeling information diffusion on internet and
in citation networks, respectively.

\paragraph{\emph{Optimization problem.}}

The unknown parameter  is the  diffusion matrix $A$, which
can be estimated by maximizing the likelihood
\begin{equation}
\begin{aligned}
& \mathop{\text{minimize}}_{\alpha_{ji}} \quad -\frac{1}{n} \sum_{c\in C^n} \log \, {\ell(  t^c;  A)}  \\
& \text{subject to} \quad \alpha_{ji} \geq 0 , j \neq i.
\label{problem_original}
\end{aligned}
\end{equation}
A nice property of the above optimization program is that it can be
further separated into $p$ independent subproblems involving
individual columns of $A$.  Specifically, the $i^{th}$ subproblem is
to infer the incoming edges into the node  $i$
\begin{equation}
\begin{aligned}
& \mathop{\text{minimize}}_{\alpha_i} \quad\, \phi(  \alpha_i)  \\
& \text{subject to} \,\,\,\,\, \alpha_{ji} \geq 0 , j \neq i,
\label{problem_original_sub}
\end{aligned}
\end{equation}
where the parameter
$\alpha_i = \{\alpha_{ji} \mid j=1, \ldots, N, j \neq i\}$ denotes the
$i^{th}$ column of $A$ and the objective function is
$$
\phi(  \alpha_i) = -\frac{1}{n}\sum_{c\in C^n} \phi_i(  t^c ;
\alpha_i),
$$
with $\phi_i(\cdot; \alpha_i)$ denoting the likelihood function for one
cascade.  For example, for the exponential transmission function, we
have
\begin{equation}
\phi_i( {t}; {\alpha}_i) = \text{log} \Bigg( \sum_{j:t_j<t_i} \alpha_{ji} \Bigg) - \sum_{j:t_j<t_i} \alpha_{ji}(t_i-t_j)
\label{g_infect}
\end{equation}
for an infected node, and
\begin{equation}
\phi_i( {t}; {\alpha}_i) = - \sum_{j:t_j<T} \alpha_{ji}(T-t_j)
\label{g_uninfect}
\end{equation}
for an uninfected node. See \cite{Gomez-Rodriguez2011Uncovering} for more details.

The problem \eqref{problem_original_sub} is convex in $ \alpha_i$ and
can be solved by a standard gradient-based algorithm.  The linear
terms in \eqref{g_infect} and \eqref{g_uninfect} act as an $\ell_1$
penalty on the unknown parameter and automatically encourage sparse
solutions.  Nonetheless, adding an explicit $\ell_1$ penalty can further
improve results. \citet{Gomez-Rodriguez2016Estimating}
propose to solve the following regularized optimization problem
\begin{equation}
  \label{original_l1}
\begin{aligned}
& \mathop{\text{minimize}}_{\alpha_i} \quad \phi(  \alpha_i) + \lambda \|  \alpha_i\|_1  \\
& \text{subject to} \quad \alpha_{ji} \geq 0 , j \neq i,
\end{aligned}
\end{equation}
using a proximal gradient algorithm \citep{parikh2014proximal}.

\subsection{Topic-sensitive Model}
\label{sec:topicsensitive}

The basic model described above makes an unrealistic assumption that
each cascade spreads based on the same diffusion matrix $A$.  However,
for example, posts on information technology usually spread much
faster than those on economy and military.  \cite{du2013uncover}
extend the basic model to incorporate this phenomena. Their
topic-sensitive model assumes that there are in total $K$ topics, and
each cascade can be represented as a topic vector in the canonical
$K$-dimensional simplex, in which each component is the weight of a
topic: $ {m}^c := (m_1^c,...,m_K^c)^{\top}$ with $\sum_k m_k^c = 1$
and $m_k^c \in [0,1]$.  Each topic $k$ is assumed to have its own
diffusion matrix $A^k= \left\{ \alpha_{ji}^k \right\}$, and the
diffusion matrix of the cascade $A^c = \left\{ \alpha_{ji}^c \right\}$
is the weighted sum of the $K$ matrices:
\begin{equation}
\alpha_{ji}^c = \sum_{k=1}^K \alpha_{ji}^k m_k^c .
\label{Le_alpha}
\end{equation}

In this way, the diffusion matrix $A^c$ can be different for different
cascades.  For each cascade $c$, the propagation model remains the
same as the basic model described in the previous section, but with the
diffusion matrix $A^c$ given in \eqref{Le_alpha}. The unknown
parameters $A^1, \ldots, A^K$ can be estimated by maximizing the
regularized log-likelihood.  \cite{du2013uncover} use a group lasso
type penalty and solve the following regularized optimization problem
\begin{equation}
  \label{Topic_Cascade_optimization_l21}
\begin{aligned}
& \mathop{\text{minimize}}_{\alpha_{ji}^k} \quad -\frac{1}{n}\sum_{c\in C^n} \phi_i\Big(  t^c ;  \big\{ \alpha_{ji}^c \big\}_{j=1}^p \Big) + \lambda \sum_j \|\alpha_{ji}\|_2  \\
& \text{subject to} \quad \alpha_{ji}^c = \sum_{k=1}^K \alpha_{ji}^k m_k^c, \\
& \qquad \qquad \quad \,\, \alpha_{ji}^k \geq 0 , \, j \neq i,
\end{aligned}
\end{equation}
with a proximal gradient based block coordinate descent algorithm.


\section{An Influence-Receptivity Based Topic-sensitive Model}\label{sec:NodeTopic}

In this section we describe our proposed influence-receptivity
model. Our motivation for proposing a new model for information
diffusion stems from the observation that the two models discussed in
Section~\ref{sec:Background} do not impose any structural assumptions
on $A$ or $A^k$ other than nonnegativity and sparsity.  However, in
real world applications we observe node-topic interactions in the
diffusion network.  For example, different social media outlets
usually focus on different topics, like information technology,
economy or military. If the main focus of a media outlet is on
information technology, then it is more likely to publish or cite news
with that topic.  Here the topics of interest of a media outlet
impart the network structure.
As another example, in a university, students may be interested in
different academic subjects, may have different music preferences, or
follow different sports. In this way it is expected that students who
share the same or similar areas of interest may have much stronger
connections.  Here the areas of interest among students impart the
structure to the diffusion network.  Finally, in the context of
epidemiology, people usually have different immune systems, and a
disease such as flu, usually tends to infect some specific people,
while leaving others uninfected.  It is very likely that the infected
people (by a specific disease) may have similar immune system, and
therefore tend to become contagious together.  Here the types of
immune system among people impart the structure.

Taking this intuition into account, we build on the topic-sensitive
diffusion model of \cite{du2013uncover} by imposing a node-topic
interaction.  This interaction corresponds to the structural
assumption on the cascade diffusion matrix $A^c$ for each cascade $c$.
As before, a cascade $c$ is represented by its weight on $K$ topics
($K \ll p$): $ m^c = (m_1^c, m_2^c, \ldots, m_K^c)^{\top} $, with
$\sum_k m_k^c = 1$ and $m_k^c \in [0,1]$.  
Each node is parameterized by its ``interest'' in each of these $K$ topics 
as two $K$ dimensional (row) vectors. 
Stacking each of these two vectors together, the ``interest'' of all the $p$
nodes form two $p \times K$ dimensional matrices.  To describe such
structure, we propose two node-topic matrices
$B_1, B_2 \in \mathbb{R}^{p \times K}$, where $B_1$ measures how much
a node can infect others (the {\it influence} matrix) and $B_2$
measures how much a node can be infected by others (the {\it
  receptivity} matrix). We use $b_{ik}^1$ and $b_{ik}^2$ to denote the
elements on $i^{th}$ row and $k^{th}$ column of $B_1$ and $B_2$,
respectively. A large $b_{ik}^1$ means that node $i$ tends to infect
others on topic $k$; while a large $b_{ik}^2$ means that node $i$
tends to be infected by others on topic $k$.  These two matrices model
the observation that, in general, the behaviors of infecting others
and being infected by others are usually different.  For example,
suppose a media outlet $i$ has many experts in a topic $k$, then it
will publish many authoritative articles on this topic. These articles
are likely to be well-cited by others and therefore it has a large
$b_{ik}^1$.  However, its $b_{ik}^2$ may not be large, because $i$ has
experts in topic $k$ and does not need to cite too many other news
outlets on topic $k$. On the other hand, if a media outlet $i$ is only
interested in topic $k$ but does not have many experts, then it will
have a small $b_{ik}^1$ and a large $b_{ik}^2$.

For a specific cascade $c$ on topic $k$, there will be an edge
$j \to i$ if and only if node $j$ tends to infect others on topic $k$
(large $b_{jk}^1$) and node $i$ tends to be infected by others on
topic $k$ (large $b_{ik}^2$).  For a cascade $c$ with the topic-weight
$ m^c$, the diffusion parameter $\alpha_{ji}^c$ is modeled as
\begin{equation}
\alpha_{ji}^c = \sum_{k=1}^K b_{jk}^1 \cdot m_k^c \cdot  b_{ik}^2.
\label{our_scalar}
\end{equation}
The diffusion matrix for a cascade $c$ can be then represented as
\begin{equation}
A^c = B_1 \cdot M^c \cdot B_2^{\top} = \sum_{k=1}^K m_k^c \cdot b_k^1 {b_k^2}^\top,
\label{our_matrix}
\end{equation}
where $M^c = \diag(m^c)$ is a diagonal matrix representing the topic weight
and $B_j = [b^j_1, \ldots, b^j_K]$
with $b^j_k$ denoting the $k^{\text{th}}$ column of $B_j$, $j=1,2$.
In a case where one does not consider self infection,
we can modify the diffusion matrix for a cascade $c$
as
\[
  A^c = B_1 M^c B_2^{\top}  - \text{diag}(B_1 M^c  B_2^{\top} ).
\]
Under the model in~\eqref{our_matrix}, the matrix $M^c$ is known for
each cascade $c \in C^n$, and the unknown parameters are $B_1$ and
$B_2$ only. The topic weights can be obtained from a topic model, such
as latent Dirichlet allocation \citep{blei2003latent}, as long as we
are given the text information of each cascade, for example, the main
text in a website or abstract/keywords of a paper. The number of
topics $K$ is user specified or can be estimated from data
\citep{hsu2016online}.  The extension to a setting with an unknown
topic distribution $M^c$ is discussed in Section \ref{sec:unknown_M}.

With a known topic distribution $M^c$, our model has $2pK$
parameters. Compared to the basic model, which has $p^2$ parameters,
and the topic-sensitive model, which has $p^2K$ parameters, we observe
that our proposed model has much fewer parameters since, usually, we
have $K \ll p$.  Based on \eqref{our_matrix}, our model can be viewed
as a special case of the topic-sensitive model where each topic
diffusion matrix $A^k$ is assumed to be of rank 1.  A natural
generalization of our model is to relax the constraint and consider
topic diffusion matrices of higher rank, which would correspond to
several influence and receptivity vectors affecting the diffusion
together.


\section{Estimation}\label{sec:Optimization}

In this section we develop an estimation procedure for parameters of
the model described in the last section.  In Section
\ref{sec:reparameterization} and \ref{sec:parameter_estimation} we
reparameterize the problem and introduce regularization terms in order
to guarantee unique solution to estimation procedure.  We then propose
efficient algorithms to solve the regularized problem in Section~\ref{sec:optimization_algorithm}.

\subsection{Reparameterization}
\label{sec:reparameterization}

The negative log-likelihood function for our model is easily obtained
by plugging the parametrization of a diffusion matrix in
\eqref{our_matrix} into the original problem \eqref{problem_original}.
Specifically, the objective function we would like to minimize is
given by
\begin{equation}
\label{eq:f_B1_B2}
f(B_1, B_2) = -\frac{1}{n} \sum_{c\in C^n} \log{\ell \big(  t^c; B_1 M^c B_2^\top \big)}.
\end{equation}
Unfortunately, this objective function is not separable in each column
of $B_1, B_2$, so we have to deal with entire matrices. Based on
\eqref{our_matrix}, recall that the diffusion matrix $A^c$ can be
viewed as a weighted sum of $K$ rank-1 matrices.  Let
$\Theta_k = b_k^1 {b_k^2}^\top$ and denote the collection of these
rank-1 matrices as $\Theta = (\Theta_1, \ldots, \Theta_K)$. With some
abuse of notation, the objective function $f(\cdot)$ in
\eqref{eq:f_B1_B2} can be rewritten as
\begin{equation}
\label{eq:f_Theta}
f(\Theta) = f(\Theta_1, \ldots, \Theta_K)
= -\frac{1}{n} \sum_{c\in C^n} \log{\ell \bigg(  t^c; \sum_{k=1}^K m_k^c \cdot \Theta_k \bigg)}.
\end{equation}
Note that since $\log \ell(\cdot)$ is convex and $A^c$ is linear
in $\Theta_k$, the objective function $f(\Theta)$ is convex in
$\Theta$ when we ignore the rank-1 constraint on $\Theta_k$.

\subsection{Parameter Estimation}
\label{sec:parameter_estimation}

To simplify the notation, we use $f(\cdot)$ to denote the objective
function in \eqref{eq:f_B1_B2} or \eqref{eq:f_Theta}, regardless of
the parameterization as $B_1,B_2$ or $\Theta$. From the
parameterization $\Theta_k = b_k^1 {b_k^2}^\top$, it is clear that if
we multiply $b_k^1$ by a constant $\gamma$ and multiply $b_k^2$ by
$1/\gamma$, the matrix $\Theta_k$ and the objective function
\eqref{eq:f_Theta} remain unchanged. In particular, we see that the
problem is not identifiable if parameterized by $B_1,B_2$. To solve
this issues we add regularization.

A reasonable and straightforward choice of regularization is the
$\ell_1$ norm regularization on $B_1$ and $B_2$. We define the
following norm
\begin{equation}
\label{eq:norm_B1_B2}
g_1(B_1, B_2) = \big\|B_1 + B_2\big\|_{1,1} \triangleq \sum_{i, k} b^1_{ik} + b^2_{ik}
\end{equation}
and the regularized objective becomes
\begin{equation}
\label{eq:f1_B1_B2}
f_1(B_1, B_2) = -\frac{1}{n} \sum_{c\in C^n} \log{\ell \big(  t^c; B_1 M^c B_2^\top \big)} + \lambda \cdot g_1(B_1, B_2),
\end{equation}
where $\lambda$ is a tuning parameter. With this regularization, if we
focus on the $k^{\text{th}}$ column, then the term we would like to
minimize is
\begin{equation}
\gamma \|b^1_k\|_1 + \frac{1}{\gamma} \|b^2_k\|_1.
\label{eq:column}
\end{equation}
Clearly, in order to minimize \eqref{eq:column} we should select
$\gamma$ such that the two terms in \eqref{eq:column} are equal. This
means that, at the optimum, the column sums of $B_1$ and $B_2$ are
equal.  We therefore avoid the scaling issue by adding the $\ell_1$
norm penalty.

An alternative choice of the regularizer is motivated by the
literature on matrix factorization \citep{jain2013low, tu2016low,
  park2016finding, ge2016matrix, zhang2017nonconvex}. In a matrix
factorization problem, the parameter matrix $X$ is assumed to be
low-rank, which can be explicitly represented as $X = UV^\top$ where
$X \in \RR^{p \times p}$, $U, V \in \RR^{p \times r}$, and $r$ is the
rank of $X$. Similar to our problem, this formulation is also not
identifiable.
The solution is to add a regularization term
$\| UU^\top - VV^\top \|_F^2$, which guarantees that the singular values of $U$ and $V$ are
the same at the optimum \citep{zhu2017global, zhang2017nonconvex,
  park2016finding, yu2020recovery}. Motivated by this approach, we
consider the following regularization term
\begin{equation}
\label{eq:guv}
g_2(B_1,B_2) = \frac 1 4 \cdot \sum_{k=1}^K \Big( \big\|b^1_k\big\|_2^2 - \big\|b^2_k \big\|_2^2 \Big)^2,
\end{equation}
which arises from viewing our problem as a matrix factorization
problem with rank-1 matrices.
The regularized objective function is therefore given by
\begin{equation}
\label{eq:f2_B1_B2}
f_2(B_1, B_2) = -\frac{1}{n} \sum_{c\in C^n} \log{\ell \big(  t^c; B_1 M^c B_2^\top \big)} + \lambda \cdot g_2(B_1,B_2).
\end{equation}
Note that for this regularization penalty, at the minimum, we have
that $g_2(B_1,B_2) = 0$ and that the $\ell_2$-norm of the columns of
$B_1$ and $B_2$ are equal. Furthermore, we can pick any positive
regularization penalty $\lambda$.

In summary, both regularizers $g_1(\cdot)$ and $g_2(\cdot)$ force the
columns of $B_1$ and $B_2$ to be balanced.  At optimum the columns
will have the same $\ell_1$ norm if $g_1$ is used and the same
$\ell_2$ norm if $g_2$ is used.  As a result, for each topic $k$, the
total magnitudes of ``influence'' and ``receptivity'' are the same. In
particular, a regularizer enforces the conservation law that the total
amount of output should be equal to the total amount of input.

The $\ell_1$ norm regularizer induces a biased sparse solution. In
contrast, the regularizer $g_2$ neither introduces bias nor encourages
a sparse solution. Since in real world applications each node is
usually interested in only a few topics, the two matrices $B_1, B_2$
are assumed to be sparse, as we state in the next section.  Taking
this into account, if the regularizer $g_2$ is used, we need to
threshold the estimator to obtain a sparse solution.  

In conclusion, the optimization problem that we are going to solve is
\begin{equation}
\begin{aligned}
&\mathop{\text{minimize}}_{B_1,B_2} \quad -\frac{1}{n} \sum_{c\in C^n} \log{\ell \big(  t^c; B_1 M^c B_2^\top \big)}  + \lambda \cdot g(B_1, B_2)   \\
& \text{subject to} \quad B_1, B_2 \geq 0,
\label{optimization}
\end{aligned}
\end{equation}
where the regularization $g(\cdot)$ is either $g_1(\cdot)$, defined in
\eqref{eq:norm_B1_B2}, or $g_2(\cdot)$, defined in \eqref{eq:guv}.

\subsection{Optimization Algorithm}
\label{sec:optimization_algorithm}

While the optimization program \eqref{problem_original} is convex in
the diffusion matrix $A$, the proposed problem \eqref{optimization} is
nonconvex in $B_1, B_2$.  Our model for a diffusion matrix
\eqref{our_matrix} is bilinear and, as a result, the problem
\eqref{optimization} is a biconvex problem in $B_1$ and $B_2$, that
is, the problem is convex in $B_1$ and $B_2$, but not jointly
convex. \cite{gorski2007biconvex} provide a survey of methods for
minimizing biconvex functions.  In general, there are no efficient
algorithms for finding the global minimum of a biconvex problem.
\cite{floudas2000deterministic} propose a global optimization
algorithm, which alternately solves primal and relaxed dual problem.
This algorithm is guaranteed to find the global minimum, but the time
complexity is usually exponential.  For our problem, we choose to
develop a gradient-based algorithm.  For the regularizer $g_1$, since
the $\ell_1$ norm is non-smooth, we develop a proximal gradient
descent algorithm \citep{parikh2014proximal}; for the regularizer
$g_2$, we use an iterative hard thresholding
algorithm \citep{yu2020recovery}.

\begin{algorithm}[tb]
   \caption{Proximal gradient descent for \eqref{optimization} with regularizer $g_1(\cdot)$}
   \label{algo_1}
\begin{algorithmic}
   \STATE {\bfseries Initialize $B_1^{(0)}$, $B_2^{(0)}$}
   \WHILE{$tolerance > \epsilon$}
   \vspace{1mm}
   \STATE $B_1^{(t+1)} = \Big[B_1^{(t)} - \eta  \nabla_{B_1} f\big(B_1^{(t)}, B_2^{(t)}\big) - \lambda \eta\Big]_+$
   \vspace{1mm}
   \STATE $B_2^{(t+1)} = \Big[B_2^{(t)} - \eta  \nabla_{B_2} f\big(B_1^{(t)}, B_2^{(t)}\big) - \lambda \eta\Big]_+$
   \ENDWHILE
\end{algorithmic}
\end{algorithm}

\begin{algorithm}[tb]
   \caption{Gradient descent with hard thresholding for \eqref{optimization} with regularizer $g_2(\cdot)$}
   \label{algo_2}
\begin{algorithmic}
   \STATE {\bfseries Initialize $B_1^{(0)}$, $B_2^{(0)}$}
   \WHILE{$tolerance > \epsilon$}
   \vspace{1mm}
   \STATE $B_1^{(t+0.5)} = \Big[B_1^{(t)} - \eta \cdot \nabla_{B_1} f\big(B_1^{(t)}, B_2^{(t)}\big) - \eta \cdot \nabla_{B_1} g_2\big(B_1^{(t)}, B_2^{(t)}\big)\Big]_+$
   \STATE $B_1^{(t+1)} = \text{Hard}\big(B_1^{(t+0.5)}, s\big)$
   \vspace{1mm}
   \STATE $B_2^{(t+0.5)} = \Big[B_2^{(t)} - \eta \cdot \nabla_{B_2} f\big(B_1^{(t)}, B_2^{(t)}\big) - \eta \cdot \nabla_{B_2} g_2\big(B_1^{(t)}, B_2^{(t)}\big)\Big]_+$
   \STATE $B_2^{(t+1)} = \text{Hard}\big(B_2^{(t+0.5)}, s\big)$
   \ENDWHILE
\end{algorithmic}
\end{algorithm}

Since the optimization problem \eqref{optimization} is nonconvex, we
need to carefully initialize the iterates $B_1^{(0)}, B_2^{(0)}$ for
both algorithms. We find the initial iterates by minimizing the
objective function $f(\Theta)$, defined in \eqref{eq:f_Theta}, without
the rank-1 constraint. As discussed earlier, the objective function
$f(\Theta)$ is convex in $\Theta$ and can be minimized by, for
example, the gradient descent algorithm. After obtaining the minimizer
$\hat \Theta = (\hat\Theta_1, \ldots, \hat\Theta_K)$, we find the best
rank-1 approximation of each $\hat\Theta_k$. According to the
Eckart-Young-Mirsky theorem, the best rank-1 approximation is obtained
by the singular value decomposition (SVD) by keeping the largest
singular value and corresponding singular vectors.  Specifically,
suppose the leading term of SVD for $\hat\Theta_k$ is denoted as
$\sigma_ku_kv_k^\top$ for each $k$, then the initial values are given
by $B_1^{(0)} = \text{Hard}\big( [u_1\sigma_1^{1/2}, \ldots, u_K\sigma_K^{1/2}] , s \big)$ and
$B_2^{(0)} = \text{Hard}\big( [v_1\sigma_1^{1/2}, \ldots, v_K\sigma_K^{1/2}] , s \big)$.
Starting from $B_1^{(0)}$, $B_2^{(0)}$, we apply one of the two
gradient-based algorithms described in Algorithm~\ref{algo_1}
and Algorithm~\ref{algo_2}, until
convergence to a pre-specified tolerance level $\epsilon$ is reached.
The
gradient $\nabla_B f(B_1, B_2)$ can be calculated by the chain rule.
The specific form depends on the transmission function used.  In
practice, the tuning parameters $\lambda$ and $s$ can be selected by
cross-validation.
Based on our
experience, both algorithms provide good estimators for $B_1$ and $B_2$.
To further accelerate the algorithm one can use the
stochastic gradient descent algorithm.


\section{Theoretical Results} \label{sec:theoretical}

In this section we establish main theoretical results.  Since the
objective function is nonconvex in $B_1, B_2$, proving theoretical
result based on the $\ell_1$ norm penalization is not
straightforward. 
For example, the usual analysis applied to nonconvex M-estimators
\citep{Loh2015Regularized} assumes a condition called restricted strong convexity, which
does not apply to our model.  Therefore, to
make headway on our problem, we focus on the optimization problem with
the regularizer $g_2$ and leverage tools that have been used in
analyzing matrix factorization problems \citep{jain2013low, tu2016low,
  park2016finding, ge2016matrix, zhang2017nonconvex, Na2019Estimating, Na2020Semiparametric}. Compared to
these works which focus on recovering one rank-$K$ matrix, our goal is
to recover $K$ rank-1 matrices.

Let $B_1^*, B_2^*$ denote the true influence and receptivity matrices;
the corresponding rank-1 matrices are given by
$\Theta_k^* = {b^1_k}^* {b^2_k}^{*\top}$, for each topic $k$. We start
by stating assumptions under which the theory is developed.
The first assumption states that the parameter matrices are sparse.

\begin{assumption}
\label{assumption:sparsity}
Each column of the true influence and receptivity matrices are assumed
to be sparse with $\|b_k^{1*}\|_0 = \|b_k^{2*}\|_0 = s^*$, where
$\|b\|_0 = \big|j: b_{j} \neq 0\big|$ denotes the number of nonzero
components of a vector.
\end{assumption}

The above assumption can be generalized in a straightforward way
to allow different columns to have different levels of sparsity.

The next assumption imposes regularity conditions on the Hessian matrix
of the objective function. First, we recall the Hessian matrix
corresponding to the objective function $\phi( \alpha)$ in
\eqref{problem_original_sub} for the basic cascade model.  For a
cascade $c$, the Hessian matrix is given by
\begin{equation}
\label{eq;original_Hessian}
\Qcal(\alpha) = D(\alpha) + X(t^c; \alpha) \cdot X(t^c; \alpha)^\top,
\end{equation}
where $D(\alpha)$ is a diagonal matrix,
\[
  X(t^c; \alpha) = h(t^c; \alpha)^{-1} \nabla_\alpha h(t^c; \alpha),
\]
with
\[
  h(t; \alpha) = \begin{cases}
    \sum_{j: t_j < t_i} H(t_i | t_j; \alpha_{ji}) &  \mbox{if } t_i < T, \\
    0 & \mbox{otherwise},
    \end{cases}
\]
and $H(t_i | t_j; \alpha_{ji})$ is the hazard function defined in
Section \ref{sec:Basicmodel}. Recalling that $\alpha \in \RR^p$
denotes the $i^{\text{th}}$ column of $A$, we have that
$\Qcal(\alpha) \in \RR^{p \times p}$.  Both $D(\alpha)$ and
$X(t^c; \alpha)$ are simple for the common transmission functions. For
example, for exponential transmission, we have that $D(\alpha) = 0$
is the all zero matrix and
\begin{equation}
\label{eq:Hessian_alpha}
\big[X(t^c; \alpha)\big]_j =
\begin{cases}
  \Big(\sum_{\ell: t_\ell < t_i} \alpha_{\ell i}\Big)^{-1} & \mbox{if } t_j < t_i \\
  0 & \mbox{otherwise.}
\end{cases}
\end{equation}
See \cite{Gomez-Rodriguez2016Estimating} for more details.

Let $[\Theta_k]_i \in \RR^{p}$ denote the $i^{\text{th}}$ column of
$\Theta_k$ and let
$\Theta^{[i]} = \Big[[\Theta_1]_i, [\Theta_2]_i, \ldots, [\Theta_K]_i
\Big] \in \RR^{p \times K}$ be the collection of $K$ such
columns. Since $A^c = \sum_{k} m_k^c \cdot \Theta_k$, we have that the
$i^{\text{th}}$ column of $A^c$ is a linear combination of
$\Theta^{[i]}$.  Therefore, the Hessian matrix of $f(\Theta)$ with
respect to $\Theta^{[i]}$ is a quadratic form of the Hessian matrices
defined in \eqref{eq;original_Hessian}.  For a specific cascade $c$,
denote the transformation matrix as
\begin{equation}
P^c =
 \begin{bmatrix}
m_1^c \cdot I_p  &  m_2^c \cdot I_p & \ldots & m_K^c \cdot I_p
 \end{bmatrix}
\in \RR^{p \times pK}.
\end{equation}
Then we have $\alpha_i^c = P^c \cdot \Theta^{[i]}$,
where $\alpha_i^c$ denotes the $i^{\text{th}}$ column of $A^c$.
Using the chain rule, we obtain that the Hessian matrix of $f(\Theta)$
with respect to $\Theta^{[i]}$ for one specific cascade $c$ is given by
\begin{equation}
H^c \big(\Theta^{[i]}\big) = {P^c}^\top \cdot \Qcal(\alpha_i^c) \cdot P^c \in \RR^{pK \times pK}.
\end{equation}
The Hessian matrix of the objective function $f(\Theta)$ with respect to $\Theta^{[i]}$
is now given as
\[
  H(\Theta^{[i]}) = \frac 1n \sum_{c} H^c(\Theta^{[i]}).
\]
We make the following assumption on the Hessian matrix.

\begin{assumption}
  \label{assumption:Hessian}
  There exist constants $\mu, L > 0$, so that
  $\mu \cdot I_{pK} \preceq H(\Theta^{[i]}) \preceq L \cdot I_{pK}$ hold uniformly
  for any $i \in \{1, \ldots, p\}$.
\end{assumption}

The optimization problem \eqref{problem_original}, used to find the
diffusion matrix $A$ for the basic cascade model, is separable across
columns of $A$ as shown in \eqref{problem_original_sub}. Similarly,
the objective function $f(\Theta)$ is separable across $\Theta^{[i]}$,
if we ignore the rank-1 constraint.  As a result, the Hessian matrix
of $f(\Theta)$ with respect to $\Theta$, is (after an appropriate
permutation of rows and columns) a block diagonal matrix in
$\RR^{p^2K \times p^2K}$ with each block given by
$H(\Theta^{[i]}) \in \RR^{p \times p}$. Therefore, Assumption
\ref{assumption:Hessian} ensures that $f(\Theta)$ is strongly convex
and smooth in $\Theta$.

The upper bound in Assumption \ref{assumption:Hessian} is easy to
satisfy. The lower bound ensures that the problem is identifiable.
The Hessian matrix depends in a non-trivial way on the network
structure, diffusion process, and the topic distributions. Without the
influence-receptivity structure, \cite{Gomez-Rodriguez2016Estimating}
establish conditions for the basic cascade model under which we can
recover the network structure consistently from the observed cascades.
The conditions require that the behavior of connected nodes are
reasonably similar among the cascades, but not deterministically
related; and also that connected nodes should get infected together
more often than non-connected nodes. Assumption
\ref{assumption:Hessian} is also related to the setting in
\cite{yu2018learning}, who consider the squared loss, where the
condition ensures that the topic distribution among the $n$ cascades
is not too highly correlated, since otherwise we cannot distinguish
them. In our setting, Assumption \ref{assumption:Hessian} is a
combination of the two cases: we require that the network structure,
diffusion process, and the topic distributions interact in a way to
make the problem is identifiable.  We refer the readers to
\cite{Gomez-Rodriguez2016Estimating} and \cite{yu2018learning} for
additional discussions.

\paragraph{\emph{Subspace distance.}} Since the factorization of $\Theta_k$ as $\Theta_k = {b^1_k} {b^2_k}^{\top}$ is not unique,
as discussed earlier, we will measure convergence of algorithms using
the subspace distance. Define the set of $r$-dimensional orthogonal
matrices as
\[
  \mathcal{O}(r) = \{ O \in \RR^{r \times r}: O^\top O = O O^\top = I_r \}.
\]
Suppose $X^* \in \RR^{p \times p}$ is a rank-$r$ matrix that can be
decomposed as $X^* = {U^*}{V^*}^\top$ with $U^*, V^* \in \RR^{p \times r}$ \
and $\sigma_i(U^*) = \sigma_i(V^*)$ where $\sigma_i(U)$ denotes the $i^{\rm{th}}$ singular value of $U$.
Let $X = U V^\top$ be an estimator of
$X^*$. The subspace distance between $X$ and $X^*$  is
measured as
\begin{equation}
\min_{O \in \mathcal{O}(r)} \Big\{ \|U - U^* O \|_F^2 + \|V - V^* O \|_F^2 \Big \}.
\end{equation}
The above formula measures the distance between matrices up to an
orthogonal rotation.  For our problem, the matrices $\Theta_k$ are
constrained to be rank-1, and the only possible rotation is given by
$o = \pm 1$.  Moreover, since $B_1, B_2 \geq 0$ are nonnegative, the
negative rotation is eliminated. As a result, the subspace distance
for our problem reduces to the usual Euclidean distance. Let
$B = [B_1, B_2]$ and $B^* = [B_1^*, B_2^*]$, then the ``subspace
distance'' between $B$ and $B^*$ is defined as
\begin{equation}
  d^2(B, B^*)
  = \min_{o_k \in \{\pm1\}}\sum_{k=1}^K  \big\|b_k^1 - {b_k^1}^*o_k\big\|_2^2 + \big\|b_k^2 - {b_k^2}^*o_k\big\|_2^2
  = \big\|B_1 - B_1^*\big\|_F^2 + \big\|B_2 - B_2^*\big\|_F^2.
\end{equation}

\paragraph{\emph{Statistical error.}} The notion of the statistical error
measures how good our estimator can be.  In a statistical estimation
problem with noisy observations, even the best estimator can only be
an approximation to the true parameter. The statistical error measures
how well the best estimator estimates the true unknown parameter.
For a general statistical estimation problem, the statistical error is
usually defined as the norm of the gradient of the objective function
evaluated at the true parameter.  For our problem, since we have
rank-1 and sparsity constraints, we define the statistical error as
\begin{equation}
\label{eq:stat_error_definition}
e_{\text{stat}} = \sup_{\Delta \in \Omega(s)} \, \big\langle \nabla_{\Theta} f(\Theta^*), \Delta \big\rangle,
\end{equation}
where the set $\Omega(s)$ is defined as
\begin{equation}
\Omega(s) = \big\{ \Delta: \Delta = [\Delta_1, \ldots, \Delta_K], \Delta_k \in \RR^{p \times p}, {\rm rank}(\Delta_k) = 2, \|\Delta_k\|_0 = 2s^2, \|\Delta\|_F = 1 \big\}.
\end{equation}
The statistical error depends on the network structure, diffusion
process, and the topic distributions, and it scales as
$n^{-1/2}$ with the sample size.

With these preliminaries, we are ready to state the main theoretical
results for our proposed algorithm. Our first result quantifies the accuracy of the initialization step. Let
\[
  \hat \Theta = \arg\min_{\Theta}\ f(\Theta)
\]
be the unconstrained minimizer of $f(\Theta)$.

\begin{theorem}
  \label{thm:initialization}
Suppose Assumption \ref{assumption:Hessian} is satisfied, and we set $s = c \cdot s^*$ in Algorithm~\ref{algo_2} for some constant $c > 1$. We have
\begin{equation}
\label{eq:init_bound}
\big\| \hat\Theta - \Theta^* \big\|_F^2 \leq \frac{2}{\mu} \big\| \nabla f(\Theta^*) \big\|_F.
\end{equation}
Furthermore,
\begin{equation}
\label{eq:bound_B0_B_star_main_text}
d^2 \big( B^{(0)}, B^* \big)
\leq \frac{80 \xi^2 K \big\| \nabla f(\Theta^*) \big\|_F^2}{\mu^2 \sigma^*},
\end{equation}
where $\xi$ is defined as $\xi^2 = 1 + \frac{2}{\sqrt{c-1}}$ and
$\sigma^* = \min_k \|\Theta_k^*\|_2$.

\end{theorem}

The upper bound obtained in \eqref{eq:init_bound} and \eqref{eq:bound_B0_B_star_main_text} can be viewed as a
statistical error for the problem without rank-1 constraints. As a
statistical error, the upper bound naturally scales with the sample
size as $n^{-1/2}$. With a large enough sample size, the initial point
will be within the radius of convergence to the true parameter such that
\begin{equation}
\label{eq:bound_current_iterate_main_text}
d^2\big( B^{(0)}, B^* \big) \leq { \frac{1}{4} \gamma \sigma^* } \cdot \min \Big\{ 1, \frac{1}{4(\mu +L )} \Big\},
\end{equation}
where $\gamma = \min\{1, \mu L/(\mu + L)\}$ .
This enables us to prove the following result.

\begin{theorem}
\label{thm:main}
Suppose Assumptions \ref{assumption:sparsity} and
\ref{assumption:Hessian} are satisfied. Furthermore, suppose the sample size $n$ is large enough
so that \eqref{eq:bound_current_iterate_main_text} holds and
\[
e_{\rm stat}^2 \leq \frac{1-\beta}{3\eta K \xi^2} \cdot \frac{\mu L }{\mu + L }\cdot  { \frac{1}{4} \gamma \sigma^* } \cdot \min \Big\{ 1, \frac{1}{4(\mu +L )} \Big\}.
\]
Then the iterates obtained by Algorithm~\ref{algo_2},
with $s = c \cdot s^*$, $c>1$, and the step size
\begin{equation}
\label{eq:step_size_selection_fixed}
\eta \leq \frac{1}{8\|B^{(0)}\|_2^2} \cdot \min\Big\{\frac{K}{2(\mu +L )}, 1\Big\},
\end{equation}
satisfy
\begin{equation}
\label{eq:linear_converge_main}
d^2 \Big( B^{(T)}, B^* \Big) \leq \beta^T \cdot d^2 \Big( B^{(0)}, B^* \Big) + \frac{C}{1-\beta}\cdot e^2_{\rm{stat}},
\end{equation}
where $\beta < 1$ and $C$ is a constant.
\end{theorem}

Theorem~\ref{thm:main} establishes convergence of iterates produced by
properly initialized Algorithm~\ref{algo_2}. The first term in
\eqref{eq:linear_converge_main} corresponds to the optimization error,
which decreases exponentially with the number of iterations, while the
second term corresponds to the unavoidable statistical error.  In
particular, Theorem~\ref{thm:main} shows linear convergence of the
iterates up to statistical error, which depends on the network
structure, diffusion process, and the topic distributions.
Note that the condition on $e_{\rm stat}$ is not stringent,
since in the case that it is not satisfied, then already
the initial point $B^{(0)}$ is accurate enough.

Proofs of Theorem \ref{thm:initialization} and \ref{thm:main} are
given in Appendix.


\section{Some Variants and Extensions}
\label{sec:Variants}

In this section we discuss several variants and application specific
extensions of the proposed model.  Section \ref{sec:friendship}
considers the extension where in addition to the influence and receptivity to topics, information propagation is further regulated by a friendship network.
Section \ref{sec:estimate_M} discusses
how we can use the $B_1$ and $B_2$ matrices to estimate the topic distribution of a new cascade for which we do not have the topic distribution apriori.  Section
\ref{sec:using_B1_B2} discusses how estimated matrices $B_1$ and $B_2$
can serve as embedding of the nodes.  Finally, in Section
\ref{sec:unknown_M} we consider estimation of $B_1, B_2$ in the setting where the
topic distributions of cascades are unknown.

\subsection{Cascades Regulated by Friendship Networks}
\label{sec:friendship}

We have used news and media outlets as our running example so far and
have assumed that each node can influence any other node. However, in
social networks, a user can only see the news or tweets published by
their friends or those she chooses to follow. If two users do not know
each other, then even if they are interested in similar topics, they
still cannot ``infect'' each others. Considering this we can modify
our model in the following way:
\begin{equation}
A^c = B_1M^cB_2^{\top}  \otimes F,
\label{F}
\end{equation}
where $\otimes$ denotes element-wise multiplication.
$F \in \{0,1\}^{p \times p}$ is a known matrix indicating whether two
nodes are ``friends'' ($f_{ji}=1$) or not ($f_{ji}=0$).  The modified
optimization problem is a straightforward extension of
\eqref{optimization} obtained by replacing the expression for $A^c$
with the new model \eqref{F}. The only thing that changes in
Algorithms~\ref{algo_1} and \ref{algo_2} is the gradient calculation.

As a further modification, we can allow for numeric values in $F$.
Here we again have $f_{ji} = 0$ if node $j$ and $i$ are not friends;
when node $j$ and $i$ are friends, the value $f_{ji} > 0$ measures how
strong the friendship is. A larger value means a stronger friendship,
and hence node $j$ could infect node $i$ in a shorter period of time.
Under this setting, we assume knowledge of whether $f_{ji}$ is 0 or not, but
 not the actual value of $f_{ji}$ when it is non-zero.
This modification is useful in dealing with information diffusion over
a social network where we know whether two nodes are friends or not,
but we do not know how strong the friendship is.  We then have to
estimate $B = [B_1, B_2]$ and $F$ jointly, resulting in a more
difficult optimization problem. A practical estimation procedure is to alternately optimize $B$ and $F$.  With a fixed $F$, the
optimization problem for $B$ can be solved using Algorithm \ref{algo_1} or
\ref{algo_2}, except for an additional element-wise multiplication
with $F$ when calculating gradient. With a fixed $B$, the optimization
problem in $F$ is convex and, therefore, can be solved by any
gradient-based iterative algorithm.

\subsection{Estimating Topic Distribution $m^c$}
\label{sec:estimate_M}

Up to now we have assumed that each topic distribution
$M^c = \diag(m^c)$ is known. However, once $B_1,B_2$ have been
estimated, we can use them to classify a new cascade $c$ by recovering
its topic-weight vector $ m^c$. For example, if an unknown disease
becomes prevalent among people, then we may be able to determine the
type of this new disease and identify the vulnerable population of
nodes.
Moreover, with estimated $B_1$ and $B_2$, we can recalculate the topic
distribution of all the cascades used to fit the model.  By comparing
the estimated distribution with the topic distribution of the cascades
we can find the ones where the two topic distributions differ a lot.
These cascades are potentially ``outliers'' or have abnormal
propagation behavior and should be further investigated.

The maximum likelihood optimization problem for estimating the topic
distribution $ m^c$ is:
\begin{equation}
\begin{aligned}
\label{eq:optimize_on_M}
& \mathop{\text{minimize}}_{m_k^c}  \quad  -\log{\ell\big(  t^c;B_1M^cB_2^{\top} \big)}  \\
& \text{subject to}  \quad \sum_k m_k^c = 1, \\
& \, \, \, \quad \qquad \qquad 0 \leq m_k^c \leq 1. \\
\end{aligned}
\end{equation}
This problem is easier to solve than \eqref{optimization} since
$A^c = B_1M^cB_2^{\top} $ is linear in $M^c$ and therefore the
problem is convex in $M^c$. The constraint $\sum_{k} m^c_k = 1$ and
$0 \leq m_k^c \leq 1$ can be incorporated in a projected gradient
descent method, where in each iteration we apply gradient descent
update on $M^c$ and project it to the simplex.

\subsection{Interpreting Node-topic Matrices $B_1$ and $B_2$}
\label{sec:using_B1_B2}

While throughout the paper we have used the diffusion of news as a
running example, our model and the notion of ``topic'' is much more
broadly applicable. As discussed before it can represent features
capturing susceptibility to diseases, as well as, geographic position,
nationality, etc. In addition to the ability to forecast future
information cascades, the influence-receptivity matrices $B_1$ and
$B_2$ can also find other uses. For example, we can use the rows of
$B_2$ to learn about the interests of users and for customer
segmentation.  In epidemiology, we can learn about the vulnerability
of population to different diseases, and allocate resources
accordingly.

The rows of $B_1,B_2$ act as a natural embedding of users in
$\mathbb{R}^{2K}$ and thus define a similarity metric, which can be
used to cluster the nodes or build recommender systems.  In
Section~\ref{sec:ExperimentsReal} illustrate how to use this embedding
to cluster and visualize nodes.  The influence-receptivity structure
is thus naturally related to graph embedding.  See
\cite{cai2018comprehensive} for a recent comprehensive survey of graph
embedding.  As a closely related work in graph embedding literature,
\cite{chen2017fast} propose a model which also embeds nodes into
$\mathbb{R}^{2K}$.  Compared to their model, our model allows for
interaction of embedding (influence and receptivity) vectors and the
topic information, resulting in more interpretable topics.  Moreover,
our model has flexibility to choose the transmission function based on
different applications and comes with theoretical results on
convergence rate and error analysis.  For example, as will be shown in
Section \ref{sec:ExperimentsReal}, for information propagation on
the internet (for example, media outlets citing articles,
Facebook and Twitter users sharing posts), we can choose the exponential transmission function; for the
citation network, the Raleigh transmission function is a more appropriate choice.

\subsection{When Topic Distribution is Unknown}
\label{sec:unknown_M}

Throughout the paper we assume that the topic distribution $M^c$ is
known for each cascade.  For example, the topic distribution can be
calculated by Topic Modeling \citep{blei2003latent} with the text
information of each cascade.  Alternatively it can come from the
knowledge of domain experts.  However, in many applications domain experts or textual information may be unavailable.  Even if
such resources are available, the topic distribution obtained from Topic
Modeling may be inaccurate or intractable in practice.  In this
case we must learn the topic distribution and the
influence-receptivity structure together.  For this problem, our
observations constitute of the timestamps for each cascade as usual, and the
variables to be optimized are $B = [B_1, B_2]$ and $M^c$ for each
cascade $c$.  A practical algorithm is to alternately optimize on $B$
and $M^c$---with a fixed
$M^c$, we follow Algorithm \ref{algo_1} or \ref{algo_2} to update $B$;
with a fixed $B$, we follow \eqref{eq:optimize_on_M} to update $M^c$
on each $c$. The two procedures are repeated until convergence.

Theoretical analysis of
this alternating minimization algorithm
under the log-likelihood in \eqref{likeli}
is beyond the scope of the paper.
For a simpler objective functions, such as the $\ell_2$ loss,
the theoretical analysis is tractable and the output of the alternating minimization algorithm
(the estimated $B$ and $M$) can be shown to converge to the true value up
to the statistical error in both $B$ and $M$. Specifically, we denote $M^*$ as the true topic
distribution and $f(\Theta, M)$ as the loss function defined in
\eqref{eq:f_Theta}.  Denote the statistical error defined in
\eqref{eq:stat_error_definition} as $e_{\text{stat}, B}$ and similarly
define the statistical error on the topic distribution $M$ as
\begin{equation}
\begin{aligned}
e_{\text{stat}, M}^2 &= \sum_{c \in C^n} \sum_{k=1}^K \Big[ \nabla_{m^c_{k}} \, f(\Theta^*, M^*) \Big]^2.
\end{aligned}
\end{equation}
Denote $B^{[t]}$ and $M^{[t]}$ as the output of the alternating
minimization algorithm at iteration $t$.  Under some additional mild
assumptions, after one iterate of the alternating minimization
algorithm we have the contraction on $B$ as
\begin{equation}
d^2 \big( B^{[t+1]}, B^* \big) \leq C_1 \cdot e^2_{{\rm{stat}}, B} + \beta_1 \cdot d^2\big(M^{[t]}, M^*\big),
\end{equation}
for some constant $C_1$ and $\beta_1 < 1$. Similarly, after one
iterate of the alternating minimization algorithm we have the
contraction on $M$ as
\begin{equation}
d^2\big(M^{[t+1]}, M^*\big) \leq C_2 \cdot e_{{\rm{stat}}, M}^2 + \beta_2 \cdot d^2 \big( B^{[t]}, B^* \big),
\end{equation}
for some constant $C_2$ and $\beta_2 < 1$.  Combining these two
inequalities, after $T$ iterations of the alternative
minimization algorithm we get
\begin{equation}
\begin{aligned}
d^2\big(B^{[T]},B^*\big) &+ d^2\big(M^{[T]}, M^*\big) \leq C_0 ( e_{{\rm stat}, M}^2 +  e_{{\rm stat}, B}^2 ) + \beta_0^T \Big[ d^2\big(B^{[0]},B^*\big) + d^2\big(M^{[0]}, M^*\big) \Big],
\end{aligned}
\end{equation}
for some constant $\beta_0 = \max\{\beta_1, \beta_2\} < 1$.  This
shows that the iterates of the alternating minimization algorithm
converge linearly to the true values up to statistical error.  We
refer the readers to Section 5 of \cite{yu2018learning} for more
details.


\section{Synthetic Data Sets} \label{sec:ExperimentsSynthetic}

In this section we demonstrate the effectiveness of our model on
synthetic data sets.  Since several existing algorithms are based on
the $\ell_1$ norm regularization, for fair comparison, we focus on our
proposed Algorithm \ref{algo_1}.\footnote{The codes are available at \url{https://github.com/ming93/Influence_Receptivity_Network}}

\subsection{Estimation Accuracy}

We first evaluate our model on a synthetic data set and compare the
predictive power of the estimated model with that of Netrate and
TopicCascade. In simulation we set $p=200$ nodes, $K = 10$ topics. We
generate the true matrices $B_1$ and $B_2$ row by row. For each row,
we randomly pick 2-3 topics and assign a random number {\sf
  Unif}$(0.8, 1.8) \cdot \zeta$, where $\zeta = 3$ with probability
0.3 and $\zeta = 1$ with probability 0.7.  We make 30\% of the values
3 times larger to capture the large variability in interests. All
other values are set to be 0 and we scale $B_1$ and $B_2$ to have the
same column sum. To generate cascades, we randomly choose a node $j$
as the source. The $j^{th}$ row of $B_1$ describes the ``topic
distribution'' of node $j$ on infecting others. Therefore we sample a
$K$ dimensional topic distribution $ m^c$ from {\sf Dir}($b^1_{j,:}$),
where $b^1_{j,:}$ is the $j^{th}$ row of $B_1$ and {\sf Dir}($\cdot$)
is Dirichlet distribution, which is widely used to generate weights 
\citep{du2013uncover, he2018xbart, glynn2019bayesian, he2020stochastic}.
According to our model \eqref{our_matrix},
the diffusion matrix of this cascade is $A^c = B_1 M^c B_2^{\top}
$. The rest of the cascade propagation follows the description in
Section \ref{sec:Basicmodel}.
For experiments we use exponential transmission function as in \eqref{eq:def_exponential_transmission}.
The diffusion process continues until
either the overall time exceeds the observation window $T = 1$, or there
are no nodes reachable from the currently infected nodes. We record
the first infection time for each node.

\begin{figure}[b!]
    \centering
    \subfigure[Negative log-likelihood on test data set]
    {
        \includegraphics[width=0.45\textwidth]{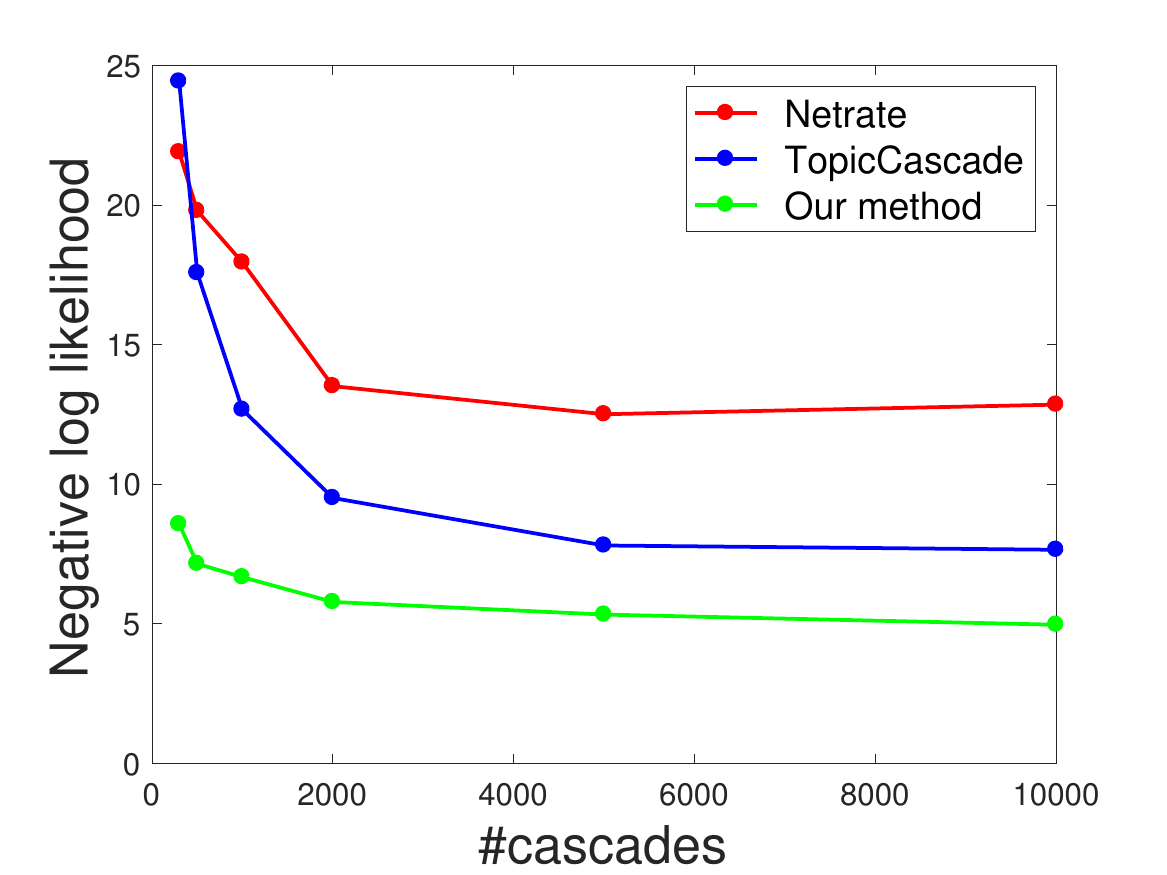}\hfill
        \label{simulation1}
    }
    \,\,\,
    \subfigure[Estimation error]
    {
        \includegraphics[width=0.45\textwidth]{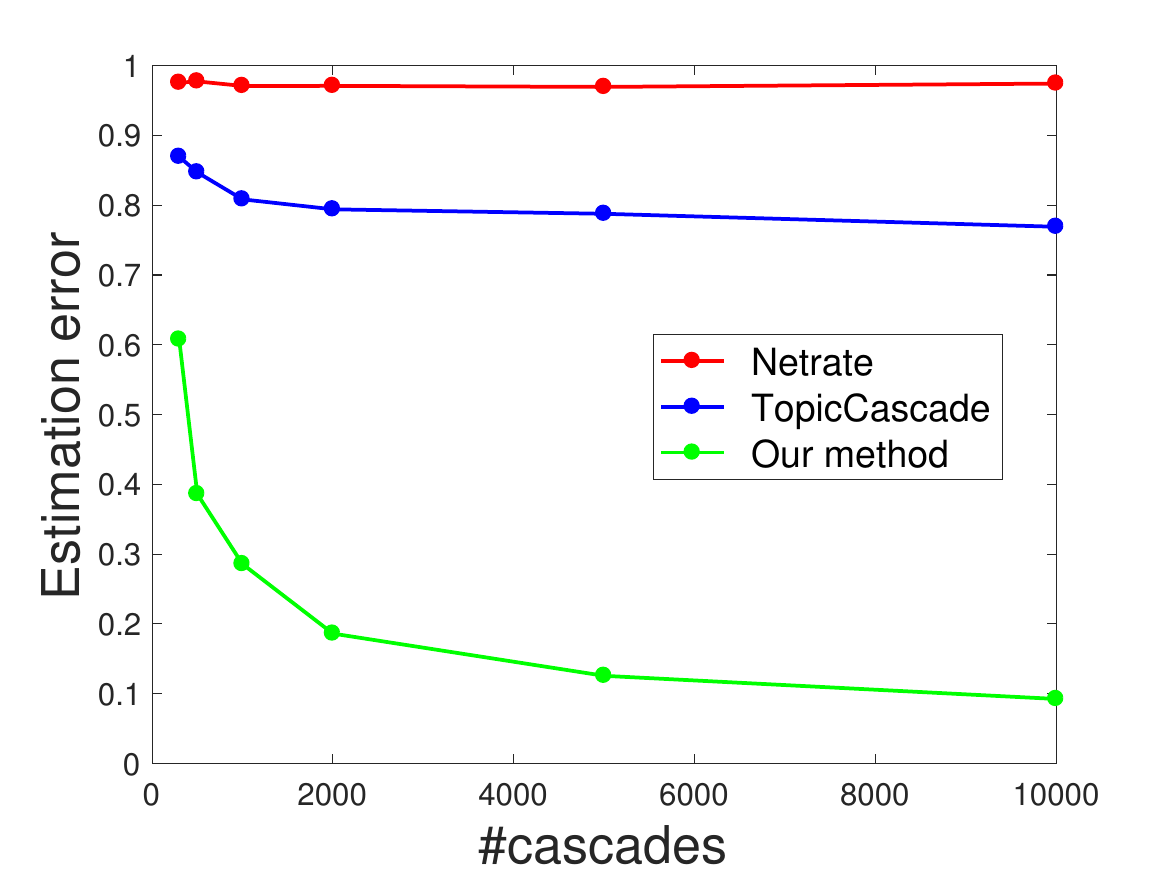}
        \label{simulation2}
    }
    \caption{Comparison of our method with Netrate and TopicCascade.}
    \label{fig:context_structures}

\end{figure}

We vary the number of cascades
$n \in \{300, 500, 1000, 2000, 5000, 10000\}$. For all three models,
we fit the model on a training data set and choose the regularization
parameter $\lambda$ on a validation data set. Each setting of $n$ is
repeated 5 times and we report the average value. We consider two
metrics to compare our model with NetRate
\citep{Gomez-Rodriguez2011Uncovering} and TopicCascade \citep{du2013uncover}:

(1) We generate independent $n = 5000$ test data and calculate
negative log-likelihood function on test data for the three models. A
good model should be able to generalize well and hence should have
small negative log-likelihood. From Figure~\ref{simulation1} we see
that, when the sample size is small, both Netrate and TopicCascade
have large negative log-likelihood on test data set; while our model
generalizes much better. When sample size increases, NetRate still has
large negative log-likelihood because it fails to consider the topic
structure; TopicCascade behaves more and more closer to our model,
which is as expected, since our model is a special case of the the
topic-sensitive model. However, our model requires substantially fewer
parameters.

(2) We calculate the true diffusion matrix $A^k$ for each topic $k$
based on our model: $A^k = B_1M_{(k)}B_2^{\top} $ where $M_{(k)}$ is
diagonal matrix with 0 on all diagonal elements but 1 on location $k$.
We also generate the estimated $\hat A^k$ from the three models as
follows: for our model we use the estimated $\hat B_1$ and $\hat B_2$;
for TopicCascade model the $\hat A^k$ is estimated directly as a
parameter of the mode; for Netrate we use the estimated $\hat A$ as
the common topic diffusion matrix for each topic $k$. Finally, we
compare the estimation error of the three models:
${\rm error} = \frac 1K \sum_{k=1}^K \frac{\|\hat A^k - A^k\|}{ \|A^k\|}$.
From Figure~\ref{simulation2} we see that both Netrate and
TopicCascade have large estimation error even if we have many samples;
while our model has much smaller estimation error.

\begin{figure}[b!]
    \centering
    \subfigure[Negative log-likelihood on test data set]
    {
        \includegraphics[width=0.45\textwidth]{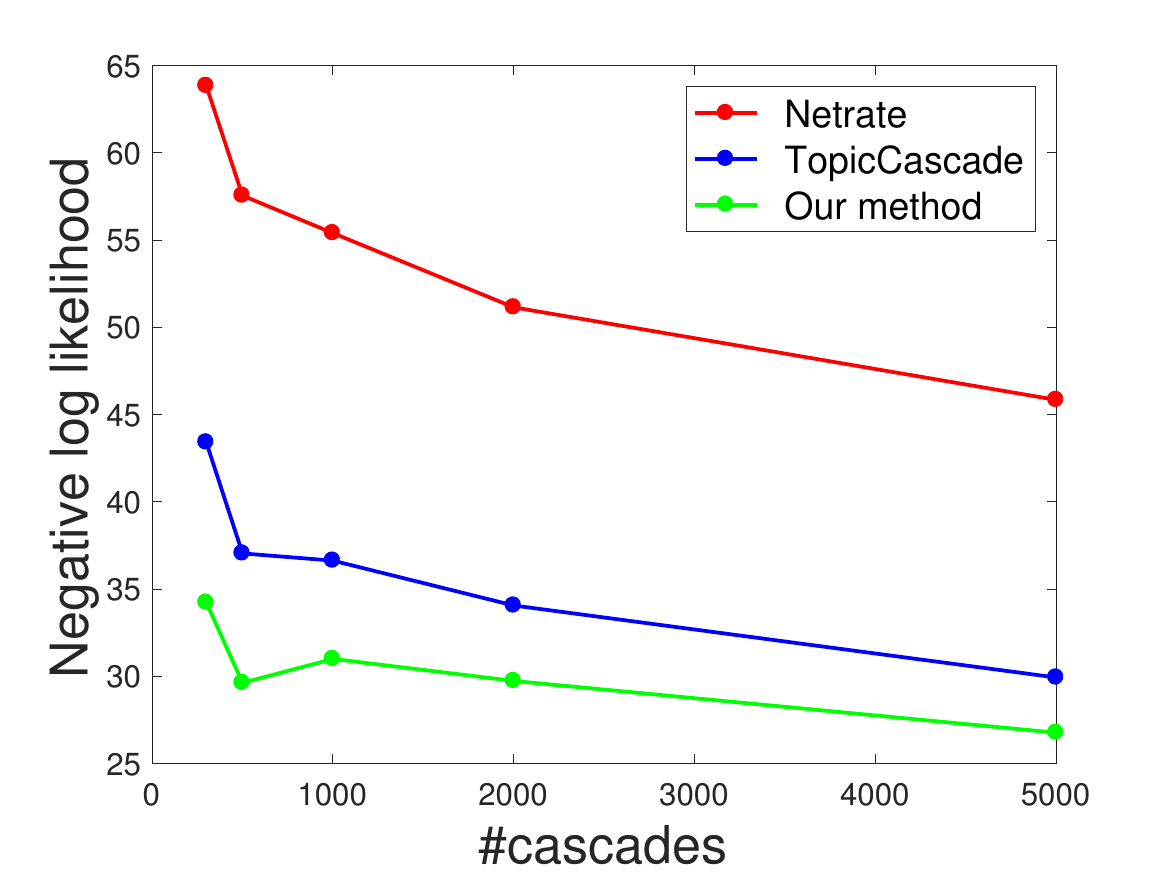}\hfill
    }
    \,\,\,
    \subfigure[Estimation error]
    {
        \includegraphics[width=0.45\textwidth]{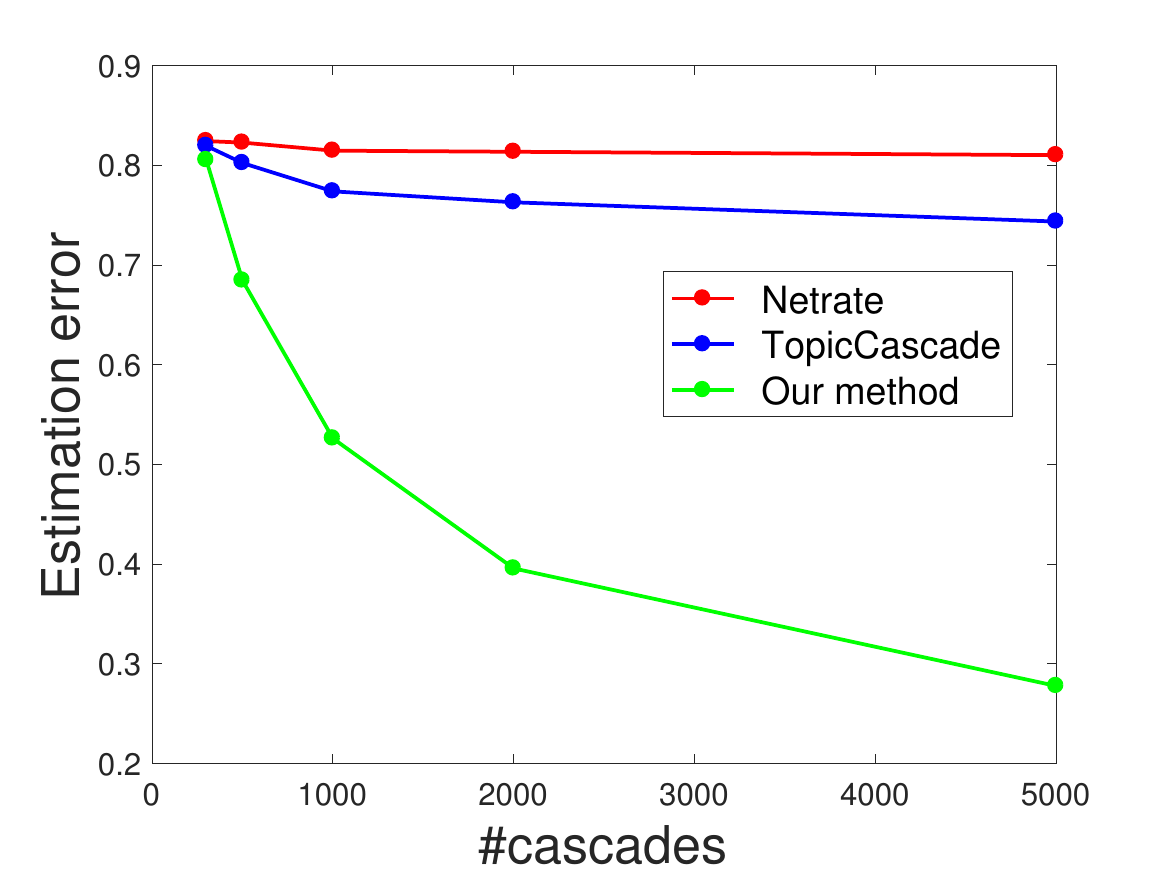}
    }
    \caption{Comparison of our method with Netrate and TopicCascade on a dense graph.}
    \label{fig:dense_graph}

\end{figure}

\paragraph{\emph{Dense graph.}} We evaluate the performance of our method on a denser graph. When generating each row of $B_1, B_2$, we randomly pick 5-6 topics instead of 2-3. This change makes infections more frequent. For many of the cascades, almost all the nodes are infected.
Since this phenomenon is not common in practice, we shrink $B_1$ and $B_2$ by half, and reduce the maximum observation time $T$ by half to make sure that infection happens across about 30\% of the nodes as before.
The comparison of our method with Netrate and TopicCascade with dense graph is shown in Figure \ref{fig:dense_graph}.
We see that the pattern is similar to the previous experiments.

\begin{figure}[b!]
    \centering
    \subfigure[Negative log-likelihood on test data set]
    {
        \includegraphics[width=0.45\textwidth]{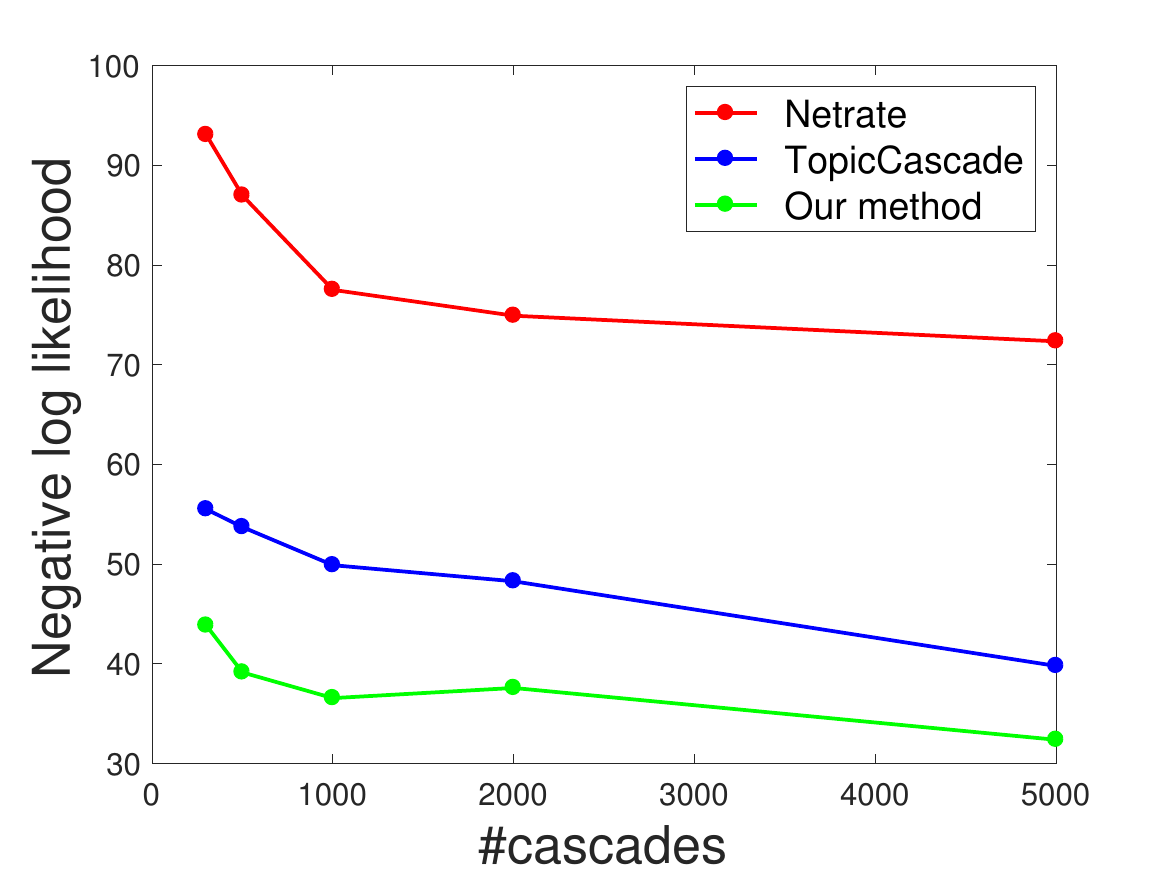}\hfill
    }
    \,\,\,
    \subfigure[Estimation error]
    {
        \includegraphics[width=0.45\textwidth]{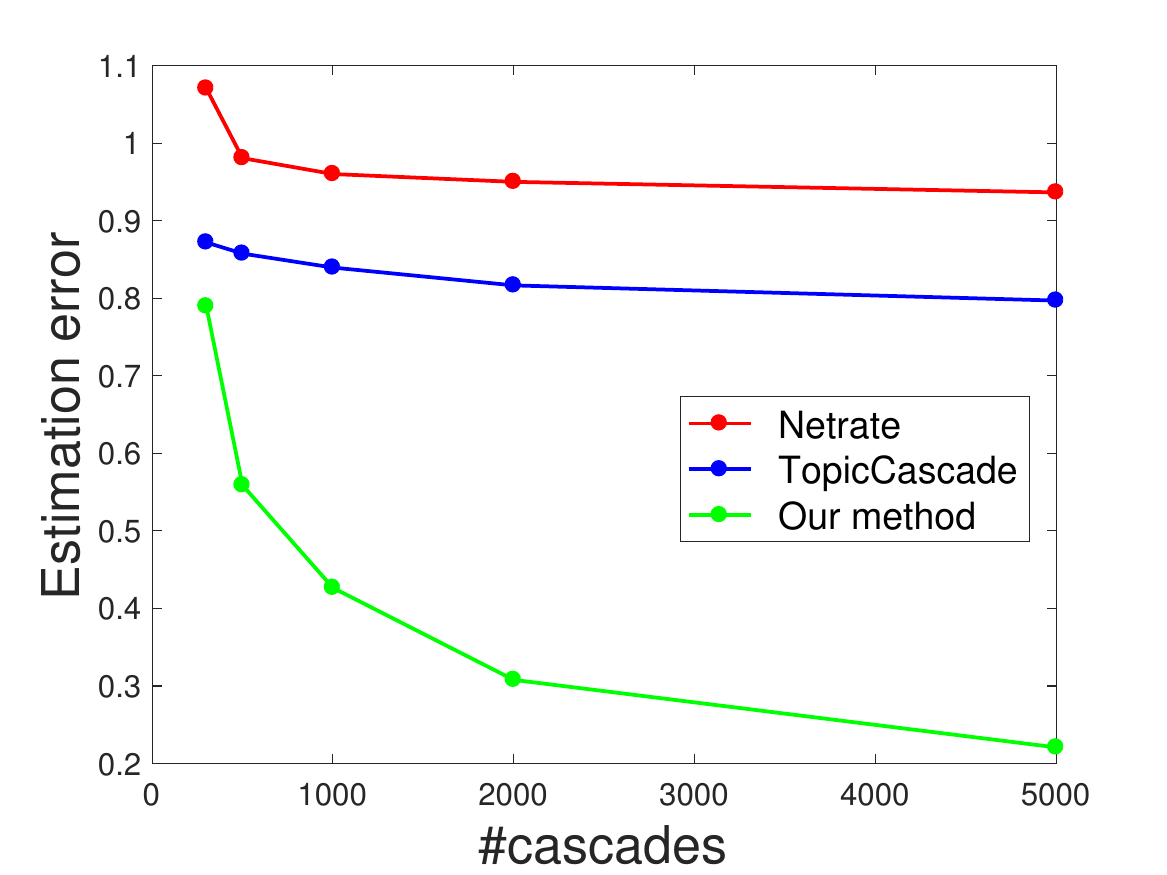}
    }
    \caption{Comparison of our method with Netrate and TopicCascade on Kronecker graph with parameter [0.8 0.6; 0.5 0.3].}
    \label{fig:Kron_1}

\end{figure}

\begin{figure}[b!]
    \centering
    \subfigure[Negative log-likelihood on test data set]
    {
        \includegraphics[width=0.45\textwidth]{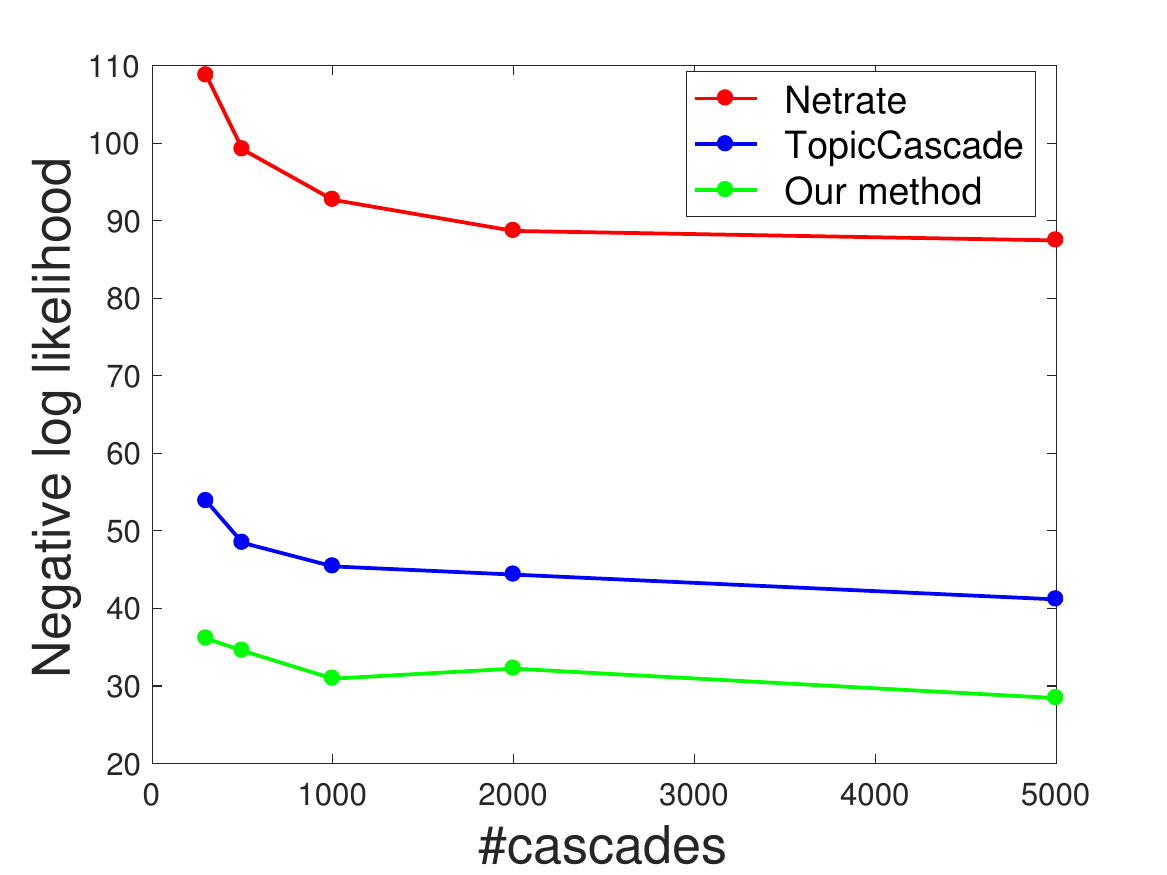}\hfill
    }
    \,\,\,
    \subfigure[Estimation error]
    {
        \includegraphics[width=0.45\textwidth]{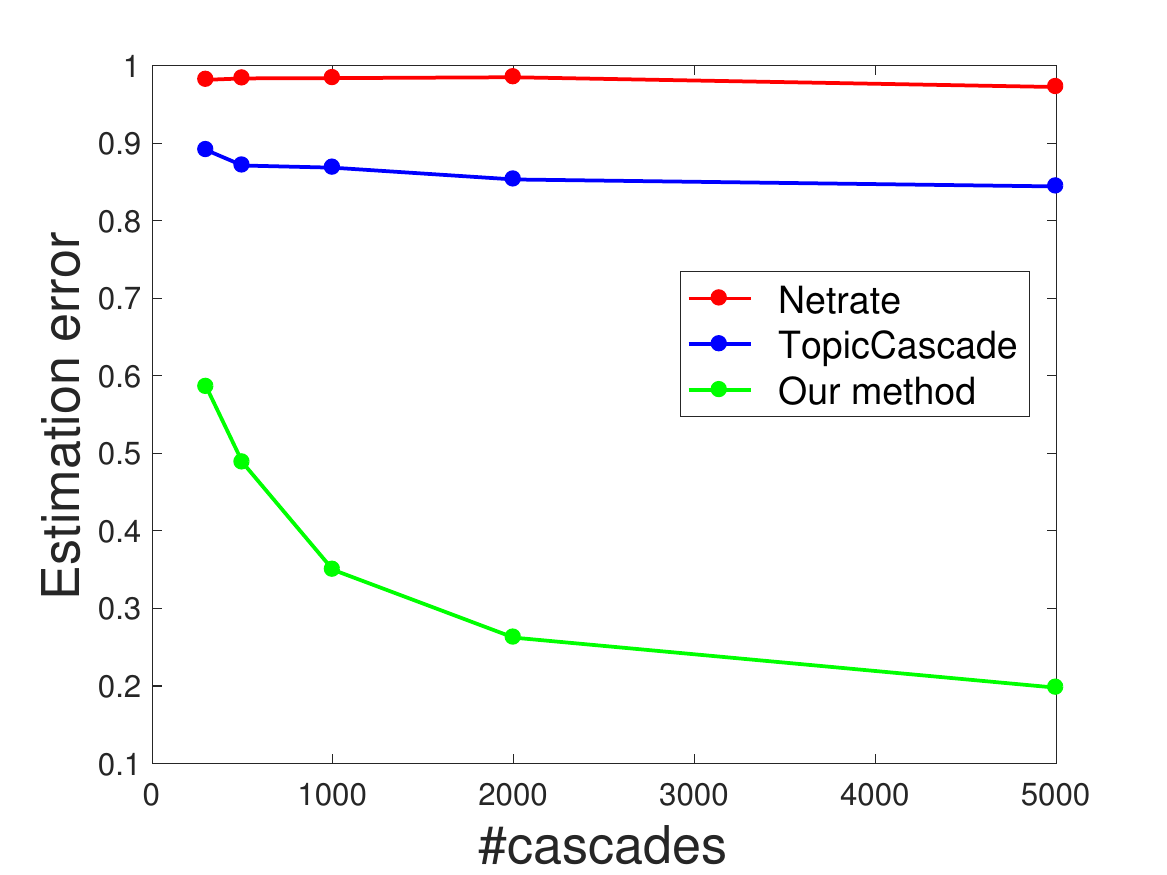}
    }
    \caption{Comparison of our method with Netrate and TopicCascade on Kronecker graph with parameter [0.7 0.7; 0.6 0.4].}
    \label{fig:Kron_2}

\end{figure}

\paragraph{\emph{Kronecker graph.}}We generate $B_1$ and $B_2$ according to the Kronecker graph \citep{leskovec2010kronecker}.
We consider two choices of parameters for generating the Kronecker graph that resemble the real world networks: the first one is [0.8 0.6; 0.5 0.3], and the second one is [0.7 0.7; 0.6 0.4].
For each choice of parameters, we follow the procedure in \cite{leskovec2010kronecker} to generate a network with $2^{11} = 2048$ nodes.
Denote this adjacency matrix as $A^{\text{Kron}} \in \RR^{2048 \times 2048}$.
Matrices $B_1, B_2 \in \RR^{2048 \times 10} $ are obtained from a non-negative matrix factorization of $A^{\text{Kron}}$, $A^{\text{Kron}} \approx B_1 B_2^\top$.
This corresponds to $K = 10$.
We randomly select $p=200$ nodes and discard others. This gives $B_1, B_2 \in \RR^{200 \times 10} $.
Finally, we zero out small values in $B_1$ and $B_2$, scale them and treat them as the true parameters
so that the percentage of infections behaves similar as the previous experiments.
Figure \ref{fig:Kron_1} and \ref{fig:Kron_2} show the comparison on Kronecker graph. Once again, our method has the best performance.

\paragraph{\emph{Compare $g_1$ and $g_2$ regularizations.}}
Both $g_1$ and $g_2$ regularizers provide good estimates for $B_1$ and $B_2$.
At optimum, the columns will have the same $\ell_1$ norm if $g_1$ is used, and the same $\ell_2$ norm if $g_2$ is used.
In simulation, the performance of using $g_1$ or $g_2$ depends on whether the true parameter has the same $\ell_1$ or $\ell_2$ column norm. In practice, the columns of $B_1$ and $B_2$ could be balanced in a much more complicated way.

\begin{figure}[b!]
    \centering
    \subfigure[Negative log-likelihood on test data set]
    {
        \includegraphics[width=0.45\textwidth]{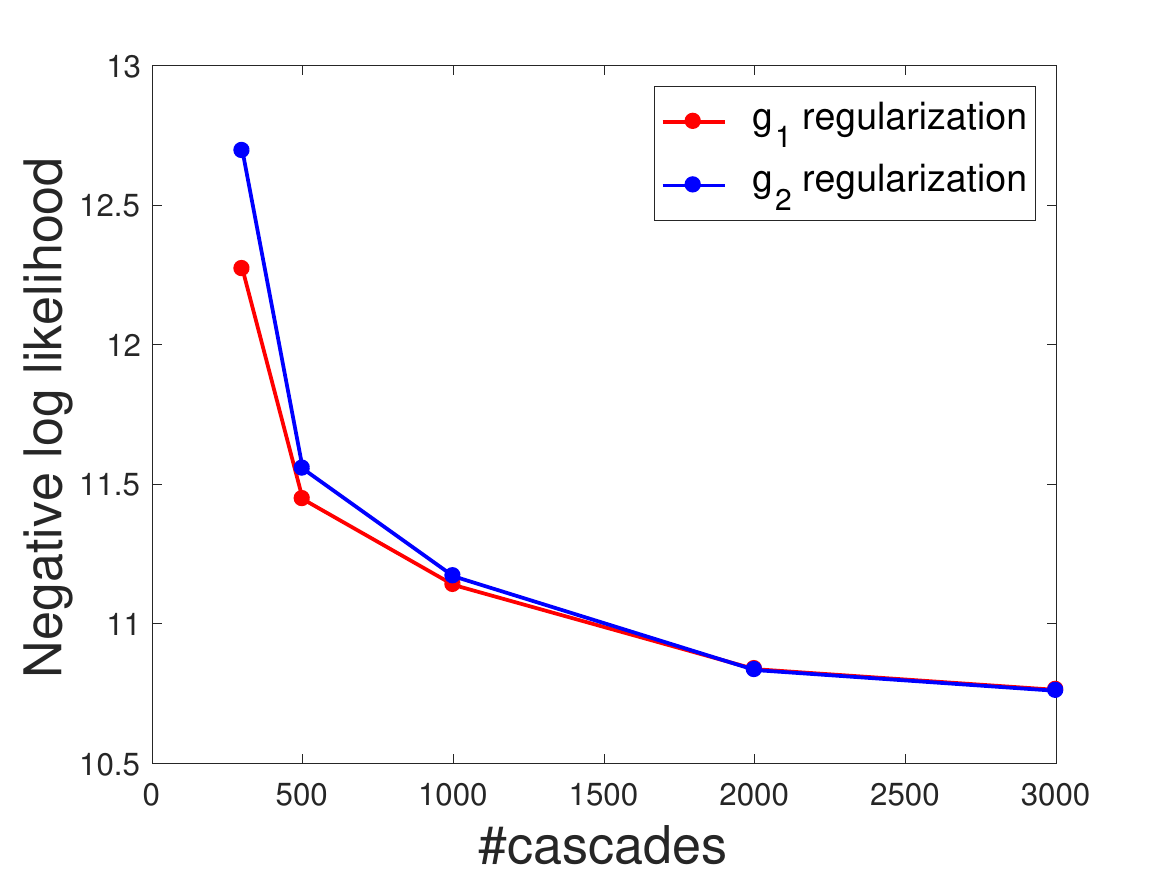}\hfill
    }
    \,\,\,
    \subfigure[Estimation error]
    {
        \includegraphics[width=0.45\textwidth]{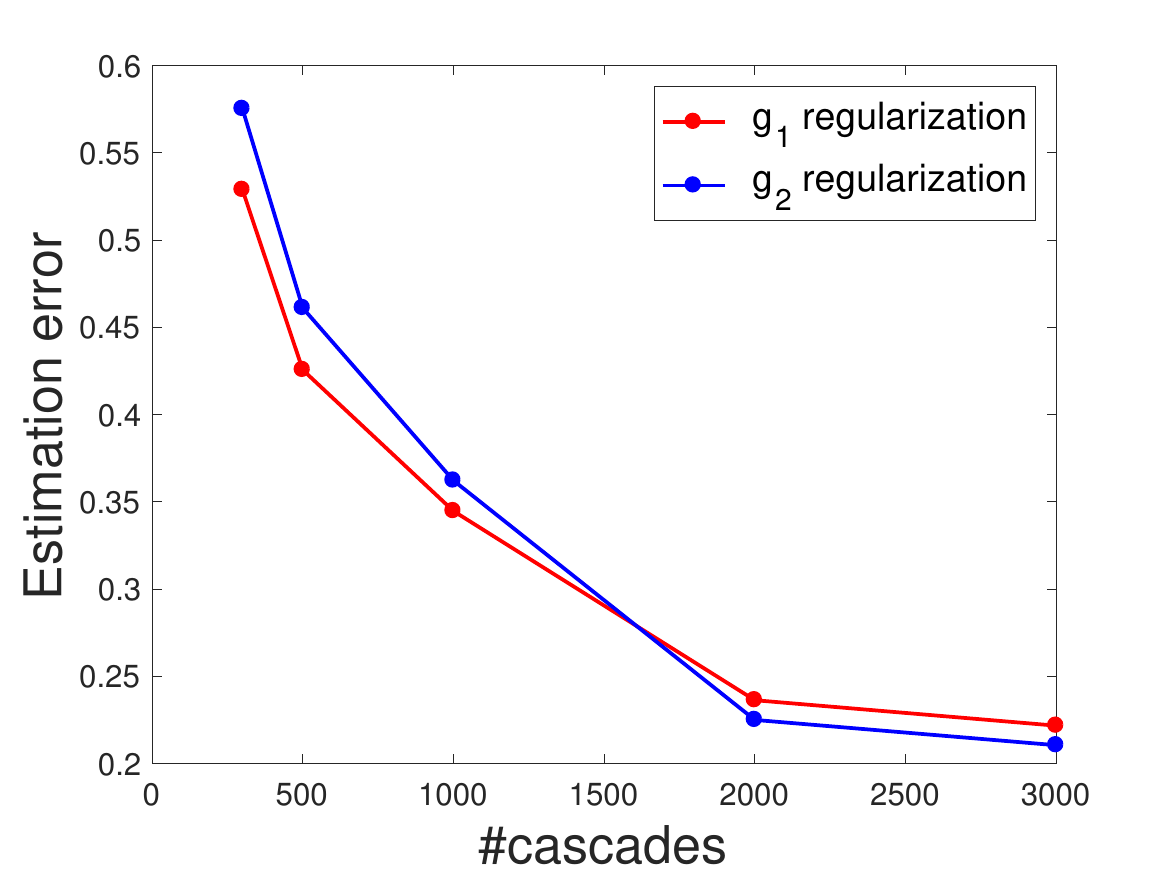}
    }
    \caption{Comparison of $g_1$ and $g_2$ regularizations.}
    \label{fig:g1_g2}

\end{figure}

For the experiment, when using $g_2$, we set $s_1 = 1.5 \cdot s_1^*$ and $s_2 = 1.5 \cdot s_2^*$ where $s_1^*$ and $s_2^*$ are the true sparsity level of $B_1$ and $B_2$; when using $g_1$, for fair comparison, we set a fixed small regularization parameter $\lambda$. To illustrate the difference between $g_1$ and $g_2$, we set $ p = 50$ and evaluate the performance of Algorithm 1 with $g_1$ and Algorithm 2 with $g_2$ on different sample sizes. We scale the true $B_1^*$ and $B_2^*$ to have the same column sum ($\ell_1$ norm).
Algorithm 2 is initialized with the solution of Algorithm 1.
Figure \ref{fig:g1_g2} shows the comparison results on different sample sizes.
We see that both methods performs well.
When sample size is small, $g_1$ seems to be slightly better, since the true values are scaled to have the same $\ell_1$ column norm.
When sample size is large, $g_2$ seems to be slightly better, since $\ell_1$ norm regularizer induces a biased solution.

\paragraph{\emph{Comparison with TopicCascade with enough samples.}}
Although our model is a special case of the topic-sensitive model, in the previous experiments, it seems like TopicCascade is not performing well even when sample size $n$ is large, especially on estimation error.
We remark that the reason is that TopicCascade has $p^2K$ parameters, while our model has only $2pK$ parameters. With $p = 200$,  TopicCascade model has 100 times more parameters than ours.
With such a large number of parameters, in order to obtain a sparse solution, we have to choose a large regularization in TopicCascade.
Such a large regularization induces a large bias on the nonzero parameters, and therefore it worsens the performance of TopicCascade.
Here we consider a lower dimensional model with $p = 10, K = 6$, and show that TopicCascade behaves similarly to our model when $n$ is large.

\begin{figure}[b!]
    \centering
    \subfigure[Negative log-likelihood on test data set]
    {
        \includegraphics[width=0.45\textwidth]{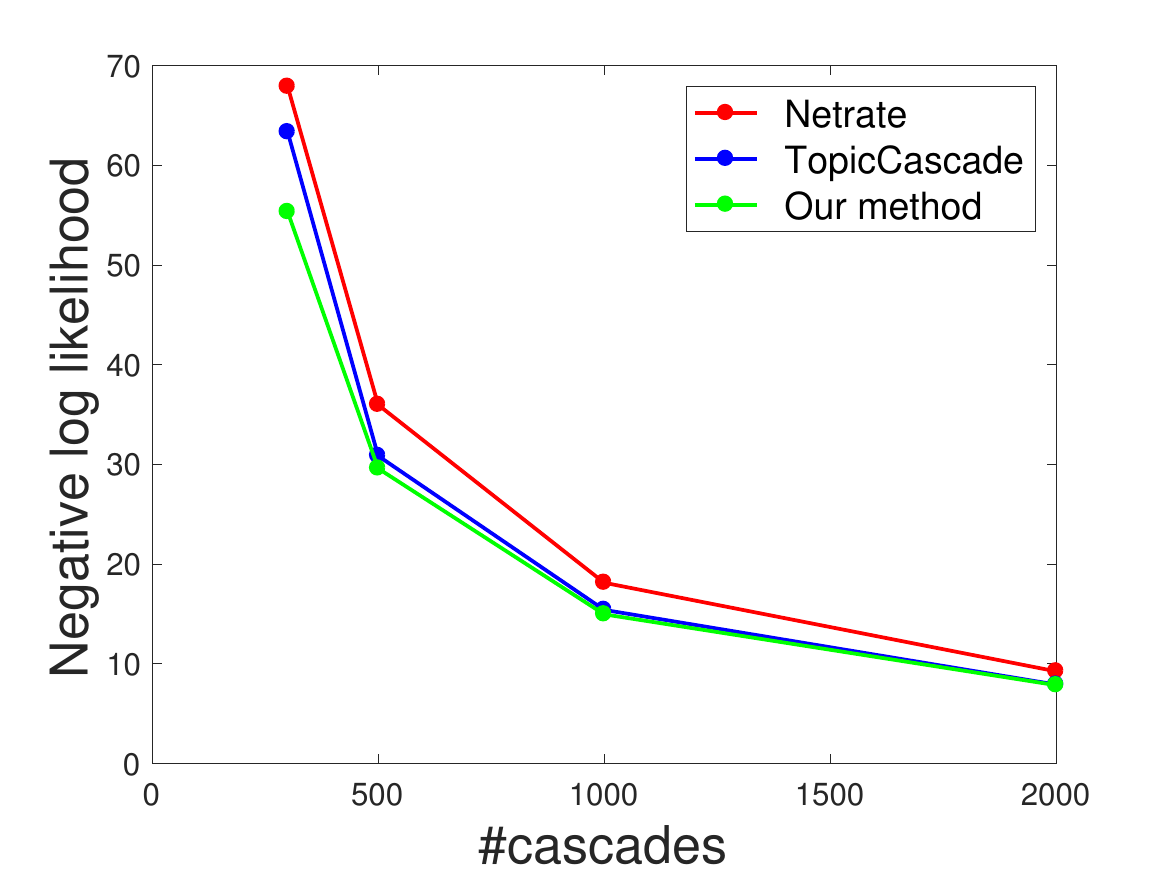}\hfill
    }
    \,\,\,
    \subfigure[Estimation error]
    {
        \includegraphics[width=0.45\textwidth]{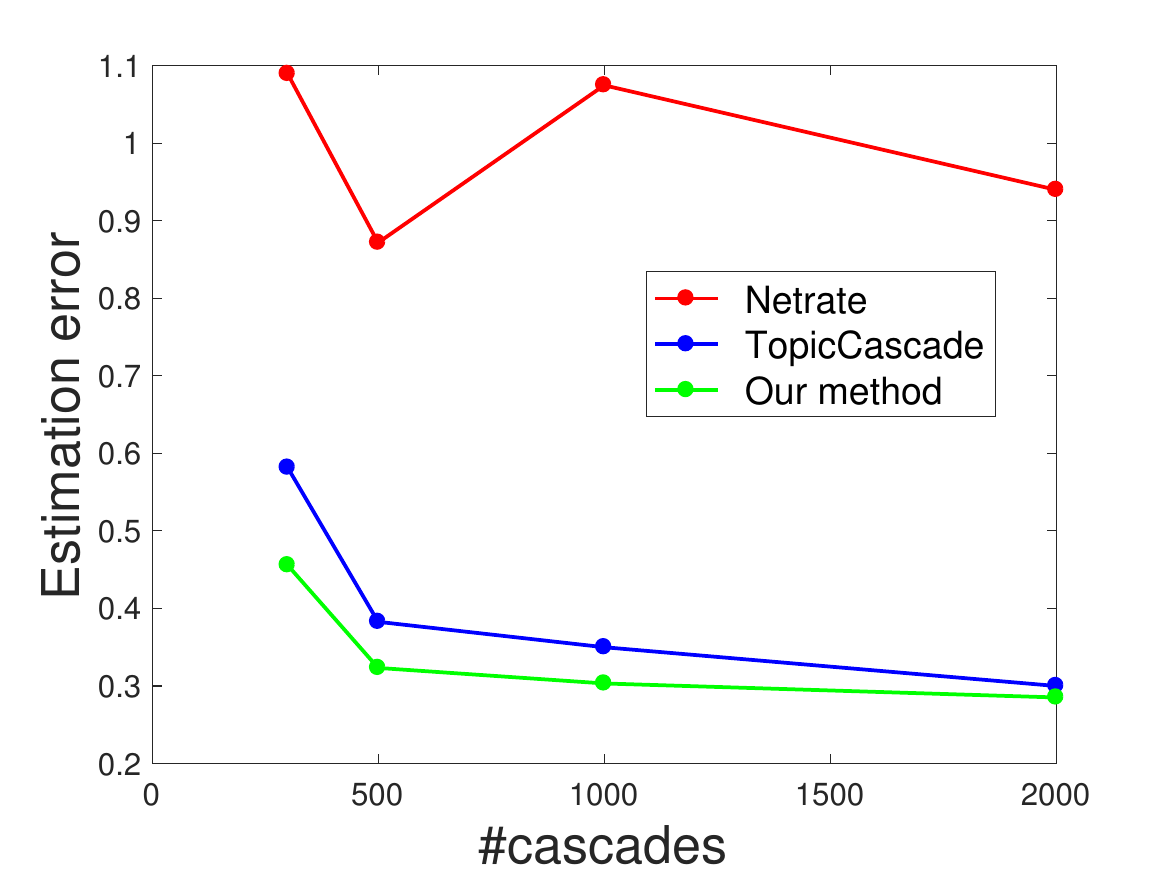}
    }
    \caption{Comparison of our method with Netrate and TopicCascade, with a small $p$.}
    \label{fig:small_p}

\end{figure}

We repeat the experiment while keeping all the other settings unchanged. Figure \ref{fig:small_p} shows the comparison of the three methods with different sample sizes.
We see that TopicCascade is almost as good as our method with large enough sample size.
We also see that Netrate performs better when $p$ and $K$ are small, in terms of negative log-likelihood.
This may be due to the small difference among topics, so one adjacency matrix suffices.
However, the performance of Netrate is still bad in terms of estimation error.
We also observe that the estimation error is not small even with small $p,K$ and large sample size.
This may be due to only a few nodes being infected in each cascade, and therefore the effective information in each cascade is low.

\paragraph{\emph{Comparison of $F_1$ score.}}
We compare the three methods using $F_1$ score.
The $F_1$ score is defined as the harmonic mean of $precision$ and $recall$: $F_1 = 2 \cdot (precision^{-1} + recall^{-1})^{-1}$, where precision is the fraction of edges in the estimated network that is also in the true network;
recall is the fraction of edges in the true network that is also in the estimated network.
Since we have $K$ topics, we calculate the $F_1$ score of each $\{ A_k \}_{k=1}^K$, and take the average.
We would like to remark that the $F_1$ score is based on the estimated discrete network, while Netrate, TopicCascade, and our model estimate continuous parameters. Therefore, the $F_1$ score is not the main focus of the comparison.
In Lasso, it is well known that one should choose a larger regularization parameter for variable selection consistency and a smaller regularization parameter for parameter estimation consistency \citep{Meinshausen2006High}.
Similarly, to obtain a better $F_1$ score, we choose a larger regularization parameter.

\begin{figure}[t]
\begin{center}
\includegraphics[width=0.6\textwidth]{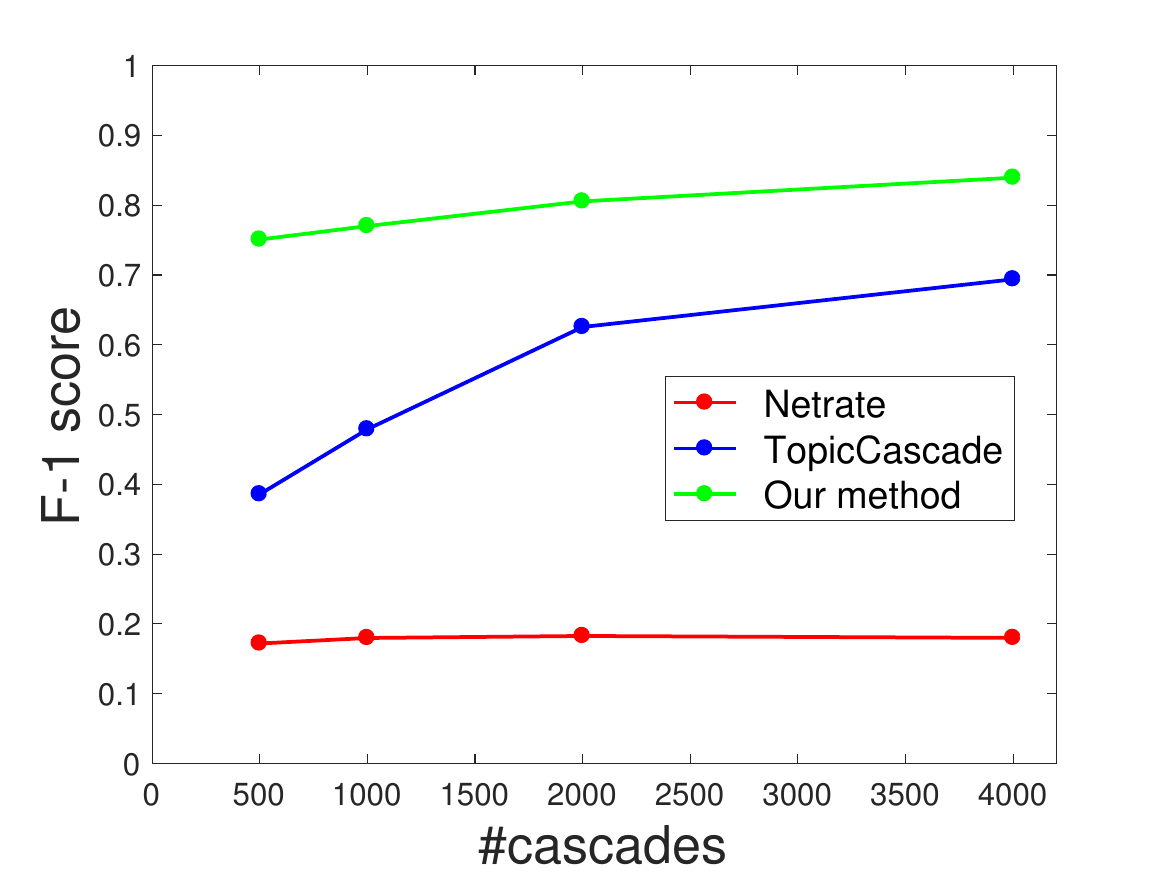}
\caption{Comparison of our method with Netrate and TopicCascade on the $F_1$ score.}
\label{fig:F1_score}
\end{center}
\end{figure}

For the experiments, we set $p = 50$ and set the regularization parameter as 20 times the optimal one selected on the validation set for parameter estimation. Figure \ref{fig:F1_score} shows the $F_1$ score of the three methods.
We see that our method has the largest $F_1$ score even when the sample size is relatively small.
With large enough sample size, both our method and TopicCascade can recover the network structure.

\subsection{Running Time}

We next compare the running times of the three methods.  For fair
comparison, for each method we set the step size, initialization,
penalty $\lambda$, and tolerance level to be the same. Also one third
of the samples are generated by each model.
For our model we follow the data generation procedure as described before;
for TopicCascade, for each topic $k$, we randomly select 5\% of the components of $A^k$ to be nonzero,
and these nonzero values are set as before as
{\sf Unif}$(0.8, 1.8) \cdot \zeta$, where $\zeta = 3$ with probability
0.3 and $\zeta = 1$ with probability 0.7;
for Netrate, we again randomly select 5\% of the components of $A$ to be nonzero with values {\sf Unif}$(0.8, 1.8) \cdot \zeta$,
and we randomly assign topic distributions.
We run the three methods on 12 kernels.  For Netrate and TopicCascade,
since they are separable in each column, we run 12 columns in
parallel; for our method, we calculate the gradient in parallel. We
use our Algorithm \ref{algo_1} for our method and the proximal
gradient algorithm for the other two methods, as suggested in
\cite{Gomez-Rodriguez2016Estimating}.  We fix a baseline model size
$n=500,p=50,K=10$, and set a free parameter $\xi$. For
$\xi = \{1,2,5,8\}$, each time we increase $n,p$ by a factor of $\xi$
and record the running time (in seconds) of each method. Table
\ref{running_time} summarizes the results based on 5 replications in
each setting.  We can see that Netrate is the fastest because it does
not consider the topic distribution. When $p$ becomes large, our
algorithm is faster than TopicCascade and is of the same order as
Netrate. This demonstrates that although our model is not separable in
each column, it can still deal with large networks.

\begin{table}[t]
\centering
\begin{tabular}{lcccc}
\hline
& $\xi = 1$ & $\xi = 2$ & $\xi = 5$ & $\xi = 8$ \\ \hline
Netrate & 1.15 & 4.42 & 53.52 & 211.0 \\
TopicCascade & 5.43 & 36.10 & 153.03 & 1310.7 \\
Our method & 9.79 & 19.83 & 91.95 & 454.9 \\\hline
\end{tabular}
\caption{Running time comparison (in sec).}
\label{running_time}
\end{table}


\section{Real World Data Set} \label{sec:ExperimentsReal}

In this section we evaluate our model on two real world data sets. We
again focus on our proposed Algorithm \ref{algo_1}.

\subsection{Memetracker Data Set} \label{sec:Memetracker}

The first data set is the MemeTracker data set
\citep{leskovec2009meme}.\footnote{Data available at
  \url{http://www.memetracker.org/data.html}} This data set contains
172 million news articles and blog posts from 1 million online sources
over a period of one year from September 1, 2008 till August 31,
2009. Since the use of hyperlinks to refer to the source of
information is relatively rare in mainstream media, the authors use
the MemeTracker methodology \citep{leskovec2016snap} to extract more
than 343 million short textual phrases. After aggregating different
textual variants of the same phrase, we consider each phrase cluster
as a separate cascade $c$. Since all documents are time stamped, a
cascade $c$ is simply a set of time-stamps when websites first
mentioned a phrase in the phrase cluster $c$. Also since the diffusion rate of information on the internet usually reaches its peak when the information first comes out and decays rapidly, we use exponential transmission function here.

For our experiments we use the top 500 media sites and blogs with the
largest 5000 cascades (phrase clusters). For each website we record
the time when they first mention a phrase in the particular phrase
cluster. We set the number of topic $K$ to be 10 as suggested in \cite{du2013uncover}, and perform Topic
Modeling (LDA) to extract 10 most popular topics.
We choose the
regularization parameter $\lambda$ based on a hold-out validation set, and
then use our Algorithm \ref{algo_1} to estimate the two node-topic
matrices. The two matrices and the key words of the 10 topics are
given in Tables~\ref{B1_meme} ($B_1$) and Table~\ref{B2_meme}
($B_2$). The keywords of the 10 topics are shown at the head of each
table; the first column is the url of the website.
Since LDA is a randomized algorithm, we run it several times and select the one that performs the best in separating the meaningful topics.
We also manually adjust the top keywords a bit by removing a few trivial words, so that they are more informative.
For example, the word ``people'' appears in several topics, and therefore we are not reporting it except for the seventh topic where ``people'' is the top-1 keyword.
The websites above
the center line in each table are the most popular websites. We have
also hand-picked some less popular websites below the center line
whose url suggest that they focus on specific topics, for example
politics, business, sports, etc. The top websites are mostly web
portals and they broadly post and cite news in many topics. Therefore
to demonstrate that our model does extract some meaningful
information, we select less popular websites below the center line and
hope we can correctly extract the topics of interest of these specific
websites.

From the two tables we can see that in general the influence matrix
$B_1$ is much sparser than the receptivity matrix $B_2$, which means
that websites tend to post news and blogs in many topics but only a
few of them will be cited by others. The websites we hand pick are not
as active as the top websites. Therefore the values for these websites
are much smaller. For the top websites we only display entries which
are above the threshold of 0.1, and leave smaller entries blank in the
two tables; for the hand selected websites, only 0 values are left
blank. From the two tables we see that our model performs quite well
on those specific websites. For example the political websites have a
large value on topic 4 (election); the business and economics websites
have large value on topic 3 (economy), etc. Those ``as expected''
large values are shown in boldface in order to highlight them.

\begin{figure}
    \centering
    \subfigure[Scatter plot of $B_1$]
    {
        \includegraphics[width=0.45\textwidth]{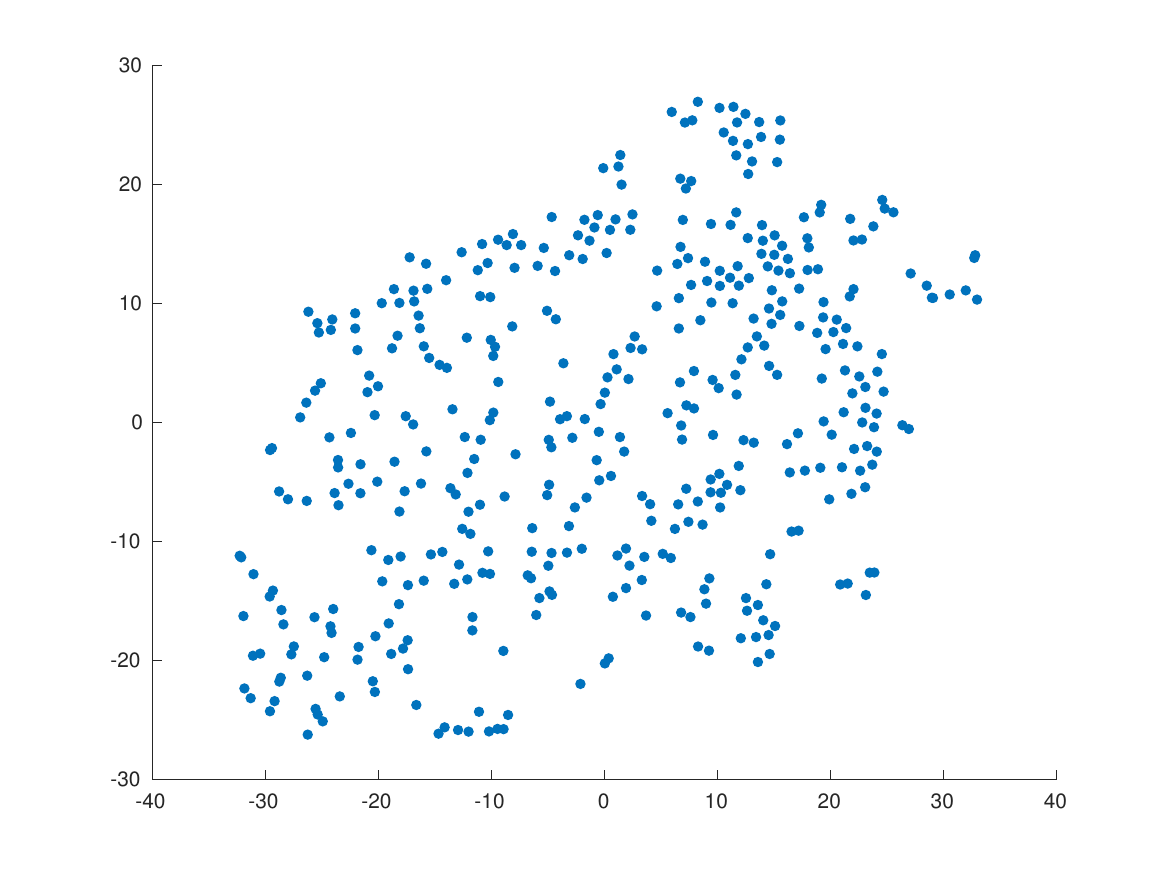}\hfill
        \label{visualization1_meme}
    }
    \,\,\,
        \subfigure[Scatter plot of $B_2$]
    {
        \includegraphics[width=0.45\textwidth]{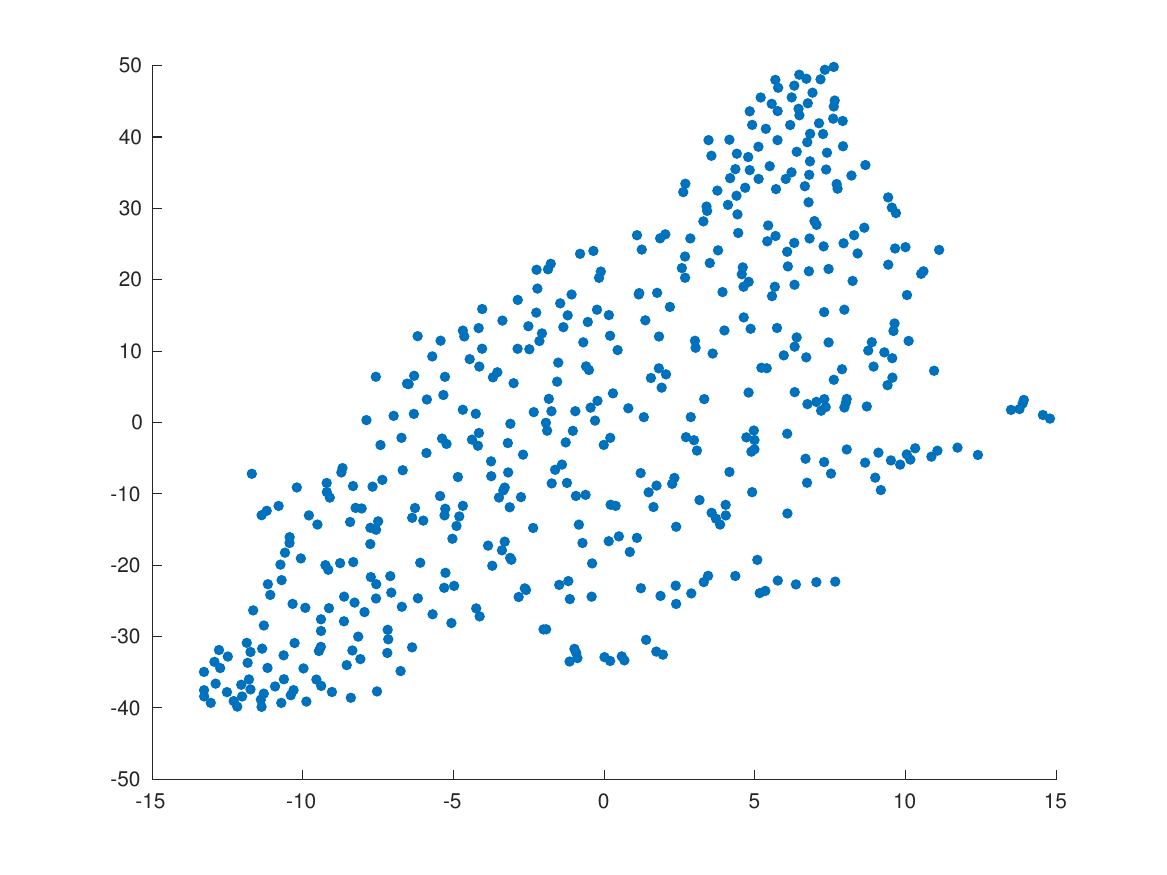}
        \label{visualization2_meme}
    }
    \caption{Scatter plot of $B_1$ and $B_2$ using t-SNE algorithm, for Memetracker data set.}
    \label{visualization_all_meme}
\end{figure}

We then visualize the estimated $B_1$ and $B_2$ using t-SNE algorithm
\citep{maaten2008visualizing} to see whether nodes are clustered with
respect to a set of topics, and whether the clusters in $B_1$
correspond to the ones in $B_2$. In $B_1$ and $B_2$, each row is a 10
dimensional vector corresponding to a website. We use t-SNE algorithm
to give each website a location in a two-dimensional map and the
scatter plot of $B_1$ and $B_2$ are given in Figure
\ref{visualization1_meme} and Figure \ref{visualization2_meme}. From
the two figures we see that these points do not form clear clusters,
which means most of the websites are in general interested in many of
the topics and they do not differ too much from each other. We can see
clearer clusters in the next example.

\begin{table}[t]
\centering
\vskip 0.15in

\begin{tabular}{lcccccc}
\hline
\\[-2.3ex]
 & train  & test & parameter &  nonzero & AIC & BIC \\\hline
 \\[-1.9ex]
Netrate & 68.5 & 81.1 & 250000 & 20143 & 2.60$\times 10^5$ & 3.65$\times 10^5$ \\
TopicCascade & 62.5& 81.8&  2500000 & 142718 & 5.08$\times 10^5$ & 1.25$\times 10^6$ \\ 
Our method & 80.3& 82.3&  10000 & 7272 &  $\bm{ 2.38\times 10^5}$ &  \bm{$2.76\times 10^5$} \\
\\[-2.3ex]
\hline
\end{tabular}

\caption{Comparison of the 3 methods on test cascades for Memetracker data set.}
\label{likeli_meme}
\end{table}

Finally we check the performance of our method on about 1500 test
cascades and compare with Netrate and TopicCascade. Since the number
of parameters are different for the three models, besides negative
log-likelihood, we also use AIC and BIC as our metrics. Table
\ref{likeli_meme} summarizes the results. The first column shows
the names of the three methods and the following columns are the
averaged negative log-likelihood on train set, averaged negative
log-likelihood on test set, number of total parameters, number of
nonzero parameters, AIC and BIC on test set calculated using the
negative log-likelihood on test set (third column) and the number of
nonzero parameters (fifth column).

From the table we see that our model has the largest negative log-likelihood on train set, and one reason for that is that our model have fewest parameters.
However, we can see that both Netrate and TopicCascade are overfitting,
while our method can generalize to test set with
little overfitting. Our method uses much fewer parameters but has
comparable negative log-likelihood on test, and also our method has
the smallest AIC and BIC value.

\begin{table}[t]
\centering

\setlength{\tabcolsep}{1.2pt}

  \resizebox{\columnwidth}{!}{%
\begin{tabular}{lcccccccccc}
\hline
\\[-2ex]
 & \begin{tabular}{@{}c@{}} energy \\ power \\ oil \\ gas \end{tabular}
 & \begin{tabular}{@{}c@{}} love \\ man \\ life \\ time \end{tabular}
 & \begin{tabular}{@{}c@{}} market \\ price \\ money \\ economy \end{tabular}
 & \begin{tabular}{@{}c@{}} obama \\ mccain \\ president \\ party \end{tabular}
 & \begin{tabular}{@{}c@{}} think \\ play \\ team \\ game \end{tabular}
 & \begin{tabular}{@{}c@{}} new \\ technology \\ system \\ data \end{tabular}
 & \begin{tabular}{@{}c@{}} people \\ clergy \\ food \\ problem \end{tabular}
 & \begin{tabular}{@{}c@{}} government \\ law \\ public \\ state \end{tabular}
 & \begin{tabular}{@{}c@{}} life \\ world \\church \\ lord \end{tabular}
 & \begin{tabular}{@{}c@{}} time \\ year \\ student \\ community \end{tabular} \\
 \\[-1.9ex]
\hline
\\[-1.9ex]
blog.myspace.com &  0.29 &  0.17 &   0.17 &   &  0.11 &  0.25 &   0.54 &   0.12 &   0.24 &  0.43\\
us.rd.yahoo.com  & 0.7 &  0.33 &   0.24 &   0.18 &  0.15 &  0.38 &   0.28 &    0.4 &   0.42 &  0.61\\
news.google.com & 0.15 &  0.13 &   0.15 &   &  0.13 &      &   &   0.15 &   &  0.65\\
startribune.com & 0.42 &  0.59 &    0.5 &    0.3 &  0.32 &  0.49 &   &   0.24 &   0.31 &     \\
news.com.au &  &  &   &   &  &  0.12 &   0.18 &   &    0.2 &     \\
breitbart.com & 0.77 &  0.47 &  &   0.15 &  0.16 &  0.37 &  &   0.25 &   0.55 &     \\
uk.news.yahoo.com & 0.51 &   0.3 &   0.36 &   &  0.17 &   0.3 &   0.33 &   0.13 &   &  0.15\\
cnn.com  &   0.13 &  0.15 &    0.5 &   0.19 &      &      &   0.34 &   &   &  0.12\\
newsmeat.com &  &      &       &   0.55 &  &      &       &       &       &     \\
washingtonpost.com  &  0.10 &  0.41 &   0.14 &  0.10 & 0.10 &  0.39 &   0.13 &   0.23 &   0.22 & \\
forum.prisonplanet.com  & 0.2 &      &  &   &  &      &  &   &   0.17 & \\
news.originalsignal.com  &  &  0.13 &   &       &  &   &   &   &  &  0.17\\
c.moreover.com  &  &  &   &   &      &  0.19 &   0.24 & &  & \\
philly.com  &  &      &   &  &  &  &   &   &   &     \\
rss.feedsportal.com&    &   0.1 &   0.14 &   &  &  0.15 &   &   0.18 &   &  0.19\\
\\[-1.9ex]
\hline
\\[-1.9ex]
foxnews.com& \bf{0.099}& 0.17& \bf{0.26}& 0.052& 0.071&&&&& 0.085\\
sports.espn.go.com&& 0.038& & 0.29& \bf{0.23}& & 0.12& 0.41& & \\
olympics.thestar.com&&&&0.013& \bf{0.036}&&&&& 0.012\\
forbes.com& \bf{0.019}&&\bf{0.028}&&0.02&&&&0.035& \\
scienceblogs.com& \bf{0.24}& 0.14& \bf{0.077}& 0.2& 0.12& \bf{0.15}& 0.092& 0.052& 0.29& 0.091\\
swamppolitics.com& &&& \bf{0.42}& 0.049&&&&& \\
cqpolitics.com&0.016& 0.23& 0.082& \bf{0.16}& 0.23& & & & 0.045&  \\
\\[-2ex]
\hline
\end{tabular}
}

\caption{ The influence matrix $B_1$ for Memetracker data set.}
\label{B1_meme}
\end{table}

\begin{table}[t]
\setlength{\tabcolsep}{1.2pt}
\resizebox{\columnwidth}{!}{%
\begin{tabular}{lcccccccccc}
\hline
\\[-2ex]
 & \begin{tabular}{@{}c@{}} energy \\ power \\ oil \\ gas \end{tabular}
 & \begin{tabular}{@{}c@{}} love \\ man \\ life \\ time \end{tabular}
 & \begin{tabular}{@{}c@{}} market \\ price \\ money \\ economy \end{tabular}
 & \begin{tabular}{@{}c@{}} obama \\ mccain \\ president \\ party \end{tabular}
 & \begin{tabular}{@{}c@{}} think \\ play \\ team \\ game \end{tabular}
 & \begin{tabular}{@{}c@{}} new \\ technology \\ system \\ data \end{tabular}
 & \begin{tabular}{@{}c@{}} people \\ clergy \\ food \\ problem \end{tabular}
 & \begin{tabular}{@{}c@{}} government \\ law \\ public \\ state \end{tabular}
 & \begin{tabular}{@{}c@{}} life \\ world \\ church \\ lord \end{tabular}
 & \begin{tabular}{@{}c@{}} time \\ year \\ student \\ community \end{tabular} \\
 \\[-1.9ex]
\hline
\\[-1.9ex]
blog.myspace.com & 0.42&  0.63&  0.28&  0.47&   0.55&  0.18&   0.29&   0.43&  0.49&   0.22\\
us.rd.yahoo.com  &  0.36&  0.28&  0.28&  0.44&   0.56&  0.19&   0.22&   0.41&  0.27&   0.18\\
news.google.com  & 0.15& &   0.10&  0.17&  & &  &   0.12&  0.11&  \\
startribune.com  & 0.19&  0.25&  0.16&  0.37&   0.38&  0.13&   0.14&   0.27&  0.23&   0.13\\
news.com.au & &   0.10& &  0.13&   0.12& &  &  & &  \\
breitbart.com  & 0.14&  0.13&  0.14&   0.3&    0.2& &  &   0.16&  0.18&  \\
uk.news.yahoo.com  & 0.12&  0.14&  0.15&  0.21&   0.14& &   0.14&   0.14&  0.13&  \\
cnn.com  & 0.12&  0.15& &  0.18&   0.16&  &  &   0.15&  0.12&  \\
newsmeat.com  &   &   &   &   &   &   &   &   &   &    \\
washingtonpost.com  & 0.12&  0.15&  0.15&  0.23&   0.17&  0.12&    0.1&   0.16&  0.18&  \\
forum.prisonplanet.com  &   &   &   & 0.10  &   &   &   &   &   0.10 &    \\
news.originalsignal.com  &  0.22&  0.23&  0.18&  0.37&   0.26& &   0.18&   0.26&  0.21&  \\
c.moreover.com  & 0.24&  0.21&  0.15&  0.37&   0.36&  0.11&   0.15&   0.34&  0.25&   0.17\\
philly.com  & 0.11&  0.15& &  0.16&   0.21& &  &   0.14&  0.11&    0.1\\
rss.feedsportal.com& &  0.11& &  0.11&  &   0.1&  &  & 0.10&   \\
\\[-1.9ex]
\hline
\\[-1.9ex]
canadianbusiness.com& 0.012& & \bf{0.061}& & & 0.017& 0.012& 0.012& &  \\
olympics.thestar.com&& 0.013& & & \bf{0.023}& & & 0.02& 0.013&  \\
tech.originalsignal.com& 0.036& 0.032& 0.04& 0.031& 0.038& \bf{0.13}& 0.037& 0.037& 0.043& 0.031\\
businessweek.com& 0.017& & \bf{0.032}& 0.012& 0.01& 0.015& 0.012& 0.012& 0.017& \\
economy-finance.com& 0.026& 0.014& \bf{0.072}& 0.024& 0.027& 0.036& & 0.03& 0.02& \\
military.com&& 0.014& & \bf{0.037}& 0.014& & & 0.02& 0.014& 0.013\\
security.itworld.com&&&&&&\bf{0.042}&&&&0.015 \\
money.canoe.ca& 0.011& & \bf{0.022}& & & 0.02& & 0.012& &  \\
computerworld.com& 0.011&&&&&\bf{0.053}&&&& \\
\\[-2ex]
\hline
\end{tabular}
}
\caption{The receptivity matrix $B_2$ for Memetracker data set.}
\label{B2_meme}

\end{table}


\subsection{Arxiv Citation Data Set}

The second data set is the ArXiv high-energy physics theory citation
network data set
\citep{leskovec2005graphs, gehrke2003overview}.\footnote{Data
  available at
  \url{http://snap.stanford.edu/data/cit-HepTh.html}} This data set
includes all papers published in ArXiv high-energy physics theory
section from 1992 to 2003. We treat each author as a node and each
publication as a cascade.  For our experiments we use the top 500
authors with the largest 5000 cascades. For each author we record the
time when they first cite a particular paper.  Since it usually takes
some time to publish papers we use Rayleigh transmission function
here.  We set the number of topic $K$ to be 6, and perform Topic
Modeling on the abstracts of each paper to extract 6 most popular
topics.  We then use our Algorithm \ref{algo_1} to estimate the two
node-topic matrices. The two matrices and the key words of the 6
topics are given in Tables~\ref{B1_citation} ($B_1$) and
Table~\ref{B2_citation} ($B_2$). Again the keywords of the 6 topics
are shown at the head of each table and the first column is the name
of the author.

We compare the learned topics to the research interests listed by the
authors in their website and we find that our model is able to
discover the research topics of the authors accurately.  For example
Arkady Tseytlin reports string theory, quantum field theory and gauge
theory; Shin'ichi Nojiri reports field theory; Burt A. Ovrut reports
gauge theory; Amihay Hanany reports string theory; Ashoke Sen reports
string theory and black holes as their research areas in their
webpages. Moreover, Ashok Das has papers in supergravity,
supersymmetry, string theory, and algebras; Ian Kogan has papers in
string theory and boundary states; Gregory Moore has papers in
algebras and non-commutativity. These are all successfully captured by
our method.

\begin{figure*}[t]
    \centering
    \subfigure[Scatter plot of $B_1$]
    {
        \includegraphics[width=0.45\textwidth]{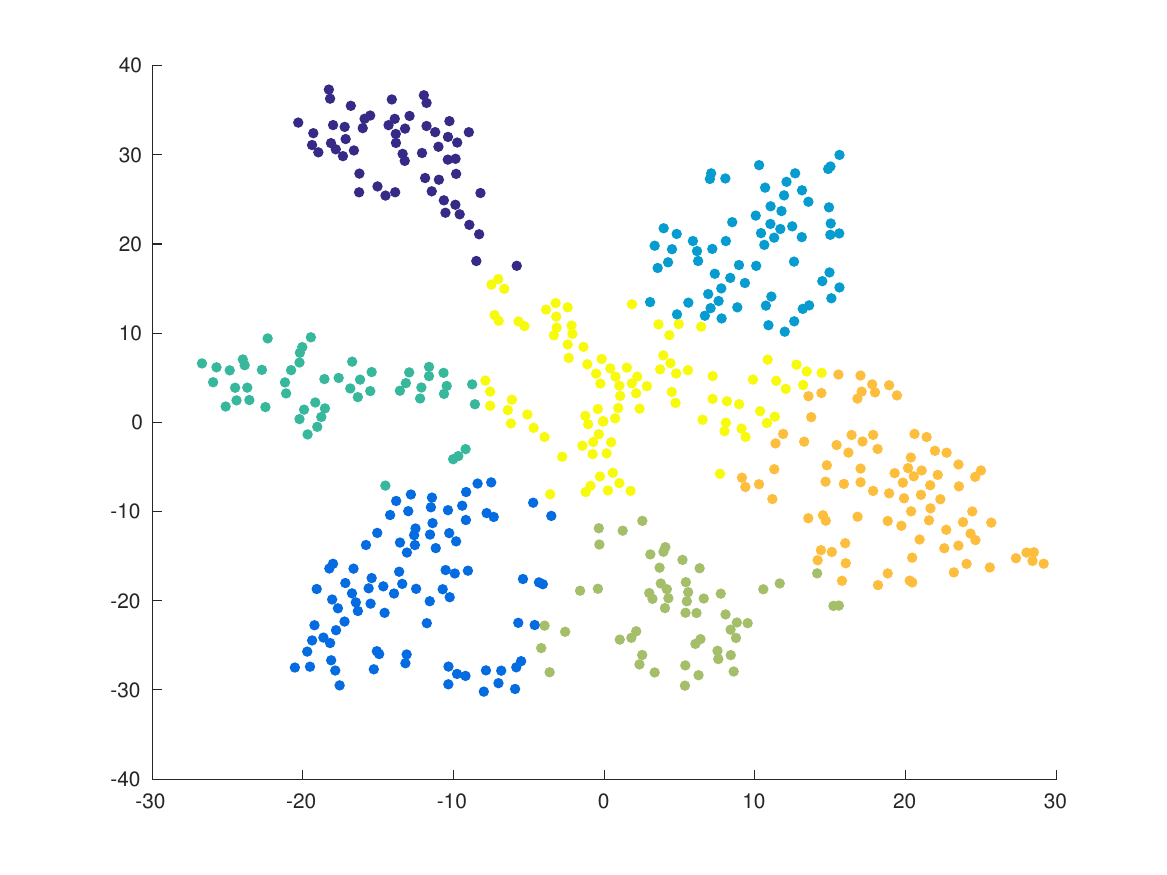}\hfill
        \label{visualization1_citation}
    }
    \,\,\,
        \subfigure[Scatter plot of $B_2$]
    {
        \includegraphics[width=0.45\textwidth]{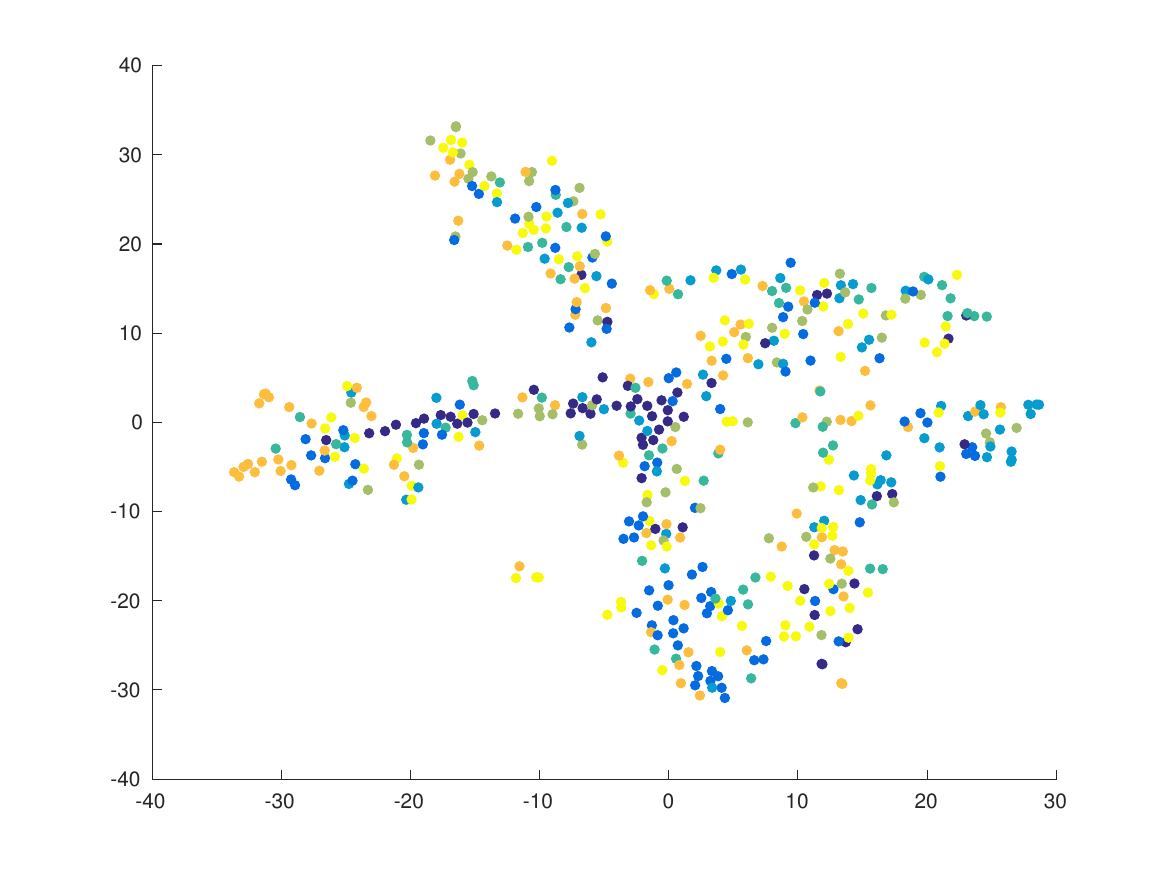}
        \label{visualization2_citation}
    }
    \caption{Scatter plot of $B_1$ and $B_2$ using t-SNE algorithm, for Citation data set.}
    \label{visualization_all_citation}
\end{figure*}

We then again visualize the estimated $B_1$ and $B_2$ using t-SNE
algorithm for which the scatter plots are shown in Figures~\ref{visualization_all_citation}.
Here we see distinct patterns in the two figures.
Figure \ref{visualization1_citation} shows 6
``petals" corresponding to the authors interested in 6 topics, while
the points in the center corresponds to the authors who have small
influence on all the 6 topics. We therefore apply $K$-Means algorithm to
get 7 clusters for the influence matrix $B_1$ as shown in Figure
\ref{visualization1_citation} (each color corresponds to one cluster),
and then plot receptivity matrix $B_2$ in Figure
\ref{visualization2_citation} using these colors. We see that although
Figure \ref{visualization2_citation} also shows several clusters, the
patterns are clearly different from Figure
\ref{visualization1_citation}. This demonstrates the necessity of
having different influence matrix $B_1$ and receptivity matrix $B_2$
in our model.

Finally we check the performance of our method on about 1200 test
cascades and compare with Netrate and TopicCascade. Table
\ref{likeli_citation} summarizes the results. Similar as before,
although Netrate and TopicCascade have smaller negative log-likelihood
on train data, our method has the best performance on test data with
significantly less parameters and little overfitting. So again we see
that our model works quite well on this citation data set.

\begin{table}[t]
\begin{center}
\begin{tabular}{lcccccc}
\hline
\\[-2.3ex]
 & train  & test & parameter &  nonzero & AIC & BIC \\\hline
 \\[-1.9ex]
Netrate & 66.8& 83.9 & 250000 & 13793 & 2.34$\times 10^5$ & 3.05$\times 10^5$ \\
TopicCascade &  67.3& 85.3&  1500000 & 57052 & 3.24$\times 10^5$ & 6.16$\times 10^5$ \\ 
Our method &78.2& 82.3&  6000 & 3738 &  \bm{$2.10\times 10^5$} &  \bm{$2.29\times 10^5$} \\
\\[-2.3ex]
\hline
\end{tabular}
\end{center}
\caption{Comparison of the 3 methods on test cascades for citation data set.}
\label{likeli_citation}
\end{table}

\begin{table}[p]
\centering
\setlength{\tabcolsep}{4pt}
\resizebox{\columnwidth}{!}{%
\begin{tabular}{lcccccc}
\hline
\\[-2ex]
 & \begin{tabular}{@{}c@{}} black \\ hole \\ energy \\ chains \end{tabular}
 & \begin{tabular}{@{}c@{}} quantum \\ model \\ field \\ theory \end{tabular}
 & \begin{tabular}{@{}c@{}} gauge \\ theory \\ field \\ effective \end{tabular}
 & \begin{tabular}{@{}c@{}} algebra \\ space \\ group \\ structure \end{tabular}
 & \begin{tabular}{@{}c@{}} states \\ space \\ noncommutative \\ boundary \end{tabular}
 & \begin{tabular}{@{}c@{}} string \\ theory \\ supergravity \\supersymmetric \end{tabular} \\
 \\[-1.9ex]
\hline
\\[-1.9ex]
Christopher N. Pope & 0.15&   0.16& 0.062&   0.12&      &     \\
Hong Lu &   0.11&   0.16& 0.067&   0.12&      &     \\
Arkady Tseytlin  &  0.019&   0.37&  0.13&   0.08&      &  0.18\\
Sergei D. Odintsov &     &   0.042&  0.29&      &      0.037& 0.013\\
Shin'ichi Nojiri &      &  0.028&  0.22&      & &     \\
Emilio Elizalde  & 0.012&  0.023&  0.11&      &   0.14&     \\
Cumrun Vafa &      &   0.17&    &   0.43&      &     \\
Edward Witten &  0.034&  0.019&     &    0.3&   0.39& 0.036\\
Ashok Das &  0.065&  0.018&     &  0.038& &  0.14\\
Sergio Ferrara &   0.41&  0.056&     &    0.2&   0.11&     \\
Renata Kallosh &   0.16&   0.49&  0.17&   0.11&  0.029&     \\
Mirjam Cvetic &     &   0.35&  0.04&  0.032&      &     0.026\\
Burt A. Ovrut &      &   0.11&  0.23&  0.083&      &     \\
Ergin Sezgin &   0.16&   0.25&  0.54&      &      &     \\
Ian Kogan  & 0.013&      &     & &   0.14&  0.11\\
Gregory Moore &      &      &     &   0.04&   0.18&     \\
I. Antoniadis &  0.21&  0.084&  0.13&   0.32&   0.07&  0.22\\
Andrew Strominger &      &   0.37&     &    0.2&      &     \\
Barton Zwiebach &  0.027& & 0.015&   0.15&    0.2&     \\
Paul Townsend &  0.036&   0.72&  0.65&   0.21&      &     \\
Robert  Myers &     &  0.075& 0.023&  0.018&      &     \\
Eric Bergshoeff & 0.096& 0.062& 0.12& 0.092&     &    \\
Amihay Hanany &     &     &    &  0.16& 0.049& 0.22\\
Ashoke Sen &   0.11&  0.15&    &  0.48&     & 0.22\\
\\[-2ex]
\hline
\end{tabular}
}
\caption{ The influence matrix $B_1$ for citation data set.}
\label{B1_citation}

\end{table}

\begin{table}[p]
\centering
\setlength{\tabcolsep}{4pt}
\resizebox{\columnwidth}{!}{%
\begin{tabular}{lcccccc}
\hline
\\[-2ex]
 & \begin{tabular}{@{}c@{}} black \\ hole \\ energy \\ chains \end{tabular}
 & \begin{tabular}{@{}c@{}} quantum \\ model \\ field \\ theory \end{tabular}
 & \begin{tabular}{@{}c@{}} gauge \\ theory \\ field \\ effective \end{tabular}
 & \begin{tabular}{@{}c@{}} algebra \\ space \\ group \\ structure \end{tabular}
 & \begin{tabular}{@{}c@{}} states \\ space \\ noncommutative \\ boundary \end{tabular}
 & \begin{tabular}{@{}c@{}} string \\ theory \\ supergravity \\supersymmetric \end{tabular} \\
 \\[-1.9ex]
\hline
\\[-1.9ex]
Christopher N. Pope &   0.5&  0.78& 0.062&   0.26&     &      \\
Hong Lu &   0.47&  0.86& 0.045&   0.25&      &      \\
Arkady Tseytlin  &  0.23&  0.88&  0.55&    0.3&   0.26&\\
Sergei D. Odintsov &     &  0.58&  0.80&  0.029&  0.14&   0.16\\
Shin'ichi Nojiri &      &  0.29&  0.35&  0.021& &   0.17\\
Emilio Elizalde  &      & 0.037&  0.18& &   0.24&  0.019\\
Cumrun Vafa & 0.098&     &     &   0.64&  0.087&   0.16\\
Edward Witten & 0.097&     &  0.29&   0.41&   0.28&    0.2\\
Ashok Das &    0.2& 0.099&  0.11&  0.023&      &   0.14\\
Sergio Ferrara &  0.51&   0.3& 0.041&   0.53&   0.13&      \\
Renata Kallosh &  0.19&   0.3&  0.58&   0.16&      &      \\
Mirjam Cvetic &  0.029&  1.4& 0.077&   0.31&      &  0.095    \\
Burt A. Ovrut & 0.021&  0.17&  0.34&   0.13& &   0.12\\
Ergin Sezgin &  0.17& 0.062&  0.38&    0.1& &      \\
Ian Kogan  &  0.061&   0.3&     &   0.05&   0.42&   0.13\\
Gregory Moore &  0.27& 0.064&  0.28&   0.51&   0.38&  0.056\\
I. Antoniadis &   0.1& 0.024& 0.042&   0.23& &    0.1\\
Andrew Strominger & 0.032&  0.58& 0.078&    0.1&  0.079&      \\
Barton Zwiebach &  0.14&     & 0.018&  0.096&  0.021&  0.068\\
Paul Townsend &  0.06&  0.12&  0.42&   0.21&      &      \\
Robert  Myers &    &  0.86&   0.2&   0.23&  0.042&   0.04\\
Eric Bergshoeff &       0.24&  0.15& 0.82& 0.27& 0.011&   \\
Amihay Hanany &    &     &    & 0.65&  0.02& 0.22\\
Ashoke Sen &    & 0.057&    & 0.16& 0.051& 0.04\\
\\[-2ex]
\hline
\end{tabular}
}

\caption{The receptivity matrix $B_2$ for citation data set.}
\label{B2_citation}

\end{table}


\section{Conclusion}
\label{sec:conclusion}

The majority of work on information diffusion has focused on
recovering the diffusion matrix while ignoring the structure among
nodes. In this paper, we propose an influence-receptivity model that
takes the structure among nodes into consideration. We develop two
efficient algorithms and prove that the iterates of the algorithm
converge linearly to the true value up to a statistical error.
Experimentally, we demonstrate that our model performs well in both
synthetic and real data, and produces a more interpretable model.

There are several interesting research threads we plan to
pursue.
In terms of modeling, an interesting future direction would be to
allow each cascade to have a different propagation rate. In our
current model, two cascades with the same topic distribution will have
the same diffusion behavior. In real world, we expect some information
to be intrinsically more interesting and hence spread much
faster. Another extension would be allowing dynamic
influence-receptivity matrices over time.
Finally, all existing work on network structure recovery from cascades
assumes that the first node observed to be infected is the source of
the diffusion. In many scenarios, the source may be latent and
directly infect many nodes. Extending our model to incorporate this
feature is work in progress.


\acks{We are extremely grateful to the associate editor, Boaz Nadler, and
two anonymous reviewers for their insightful comments that helped improve
this paper. This work is partially supported by an IBM Corporation Faculty
Research Fund
and the William S. Fishman Faculty Research Fund
at the University of Chicago Booth School of
Business. This work was completed in part with resources provided by
the University of Chicago Research Computing Center.}




\newpage

\appendix
\section{Technical proofs}

\subsection{Proof of Theorem \ref{thm:initialization}.}

Since $f(\Theta)$ is strongly convex in $\Theta$, we have
\begin{equation}
f(\hat\Theta) - f(\Theta^*) - \big\langle \nabla f(\Theta^*), \hat\Theta - \Theta^* \big\rangle \geq \frac{\mu}{2} \big\| \hat\Theta - \Theta^* \big\|_F^2.
\end{equation}
On the other hand, since $\hat\Theta$ is the global minimum, we have
\begin{equation}
f(\hat\Theta) \leq f(\Theta^*).
\end{equation}
Combining the above two inequalities, we obtain
\begin{equation}
\frac{\mu}{2}\big\| \hat\Theta - \Theta^* \big\|_F^2 \leq - \big\langle \nabla f(\Theta^*), \hat\Theta - \Theta^* \big\rangle \leq \big\| \nabla f(\Theta^*) \big\|_F \cdot \big\| \hat\Theta - \Theta^* \big\|_F
\end{equation}
and
\begin{equation}
\big\| \hat\Theta - \Theta^* \big\|_F \leq \frac{2}{\mu} \big\| \nabla f(\Theta^*) \big\|_F.
\end{equation}
This shows that for any $k$, we have
\begin{equation}
\big\| \hat\Theta_k - \Theta_k^* \big\|_F \leq \frac{2}{\mu} \big\| \nabla f(\Theta^*) \big\|_F.
\end{equation}
According to the construction of the initialization point, the rank-1 SVD of
$\Theta_k$ is given by $\sigma_k u_k v_k^\top$. Since it is the best
rank-1 approximation of $\hat \Theta_k$, we have that
\begin{equation}
\big\| \sigma_k u_k v_k^\top - \hat\Theta_k \big\|_F \leq  \big\| \hat\Theta_k - \Theta_k^* \big\|_F.
\end{equation}
By the triangular inequality
\begin{equation}
\big\| \sigma_k u_k v_k^\top - \Theta_k^* \big\|_F \leq
\big\| \sigma_k u_k v_k^\top - \hat\Theta_k \big\|_F + \big\| \hat\Theta_k - \Theta_k^* \big\|_F \leq 2 \big\| \hat\Theta_k - \Theta_k^* \big\|_F \leq \frac{4}{\mu} \big\| \nabla f(\Theta^*) \big\|_F.
\end{equation}
Then by Lemma 5.14 in \cite{tu2016low} we have
\begin{equation}
\big\| {b_k^1}^{(0)} - {b_k^1}^* \big\|_2^2 + \big\| {b_k^2}^{(0)} - {b_k^2}^* \big\|_2^2 \leq \frac{2}{\sqrt 2 - 1} \cdot \frac{\big\| \sigma_k u_k v_k^\top - \Theta_k^* \big\|_F^2}{\|\Theta_k^*\|_2}.
\end{equation}
Let $\sigma^* = \min_k \|\Theta_k^*\|_2$. Using Lemma 3.3 in
\cite{li2016stochastic}, we have the following upper bound on the
initialization $B^{(0)} = \big[ B_1^{(0)}, B_2^{(0)} \big]$,
\begin{equation}
\label{eq:bound_B0_B_star}
d^2 \big( B^{(0)}, B^* \big) \leq \xi^2 \cdot \frac{2K}{\sqrt 2 - 1} \cdot \frac{16 \big\| \nabla f(\Theta^*) \big\|_F^2}{\mu^2 \sigma^*}
\leq \frac{80 \xi^2 K \big\| \nabla f(\Theta^*) \big\|_F^2}{\mu^2 \sigma^*},
\end{equation}
where $\xi$ is defined as $\xi^2 = 1 + \frac{2}{\sqrt{c-1}}$ with $c$
set as $s = cs^*$ as in Theorem \ref{thm:main}.

\subsection{Proof of Theorem \ref{thm:main}.}

The key part of the proof is to quantify the estimation error after
one iteration. We then iteratively apply this error bound.  For
notation simplicity, we omit the superscript indicating the iteration
number $t$ when quantifying the iteration error. We denote the current
iterate as $B = [B_1, B_2]$ and the next iterate as
$B^+ = [B_1^+, B_2^+]$.  Recall that the true values are given by
$B^* = [B_1^*, B_2^*]$ with columns given by ${b_k^1}^*, {b_k^2}^*$.
The $k^{\text{th}}$ columns of $B_1, B_2, B_1^+, B_2^+$ are denoted as $b_k^1, b_k^2, b_k^{1+}, b_k^{2+}$.
We use $b_k$ and $b_k^+$ to denote $b_k = [b_k^1, b_k^2]$ and $b_k^+ = [b_k^{1+}, b_k^{2+}]$.

According to the update rule given in Algorithm \ref{algo_2}, we have
\begin{align}
B_1^+ & = \text{Hard} \Big( B_1 - \eta \cdot \nabla_{B_1} f\big(B_1, B_2\big) - \eta \cdot \nabla_{B_1} g\big(B_1, B_2\big), s \Big), \\
B_2^+ & = \text{Hard} \Big( B_2 - \eta \cdot \nabla_{B_2} f\big(B_1, B_2\big) - \eta \cdot \nabla_{B_2} g\big(B_1, B_2\big), s \Big),
\end{align}
with the regularization term
$ g(B_1,B_2) = \frac 1 4 \cdot \sum_{k=1}^K \Big( \big\|b^1_k\big\|_2^2 - \big\|b^2_k \big\|_2^2 \Big)^2 $ given in
\eqref{eq:guv}.
Note that, since the true values $B_1^*, B_2^*$ are nonnegative and
the negative values only make the estimation accuracy worse, we can
safely ignore the operation $[B]_+$ in the theoretical
analysis. Moreover, when quantifying the estimation error after one
iteration, we assume that the current estimate $B$ is not too far away
from the true value $B^*$ in that
\begin{equation}
\label{eq:bound_current_iterate}
d^2(B,B^*) \leq { \frac{1}{4} \gamma \sigma^* } \cdot \min \Big\{ 1, \frac{1}{4(\mu +L )} \Big\},
\end{equation}
where $\gamma = \min\{1, \mu L/(\mu + L)\}$ and
$\sigma^* = \min_k \|\Theta_k^*\|_2$.
This upper bound \eqref{eq:bound_current_iterate} is satisfied for $B^{(0)}$ when the sample size is large enough, as assumed in \eqref{eq:bound_current_iterate_main_text}.
In the proof, we will show that
\eqref{eq:bound_current_iterate} is also satisfied in each iteration
of Algorithm \ref{algo_2}. Therefore we can recursively apply the
estimation error bound for one iteration.

Let
\[
  S_1 = \text{supp}(B_1) \cup \text{supp}(B_1^+) \cup \text{supp}(B_1^*)
 ~\text{ and }~
  S_2 = \text{supp}(B_2) \cup \text{supp}(B_2^+) \cup \text{supp}(B_2^*)
\]
denote the nonzero positions of the current iterate, next iterate, and the true value.
Similarly, let
\[
  S_{1k} = \text{supp}(b^1_k) \cup \text{supp}(b_k^{1+}) \cup \text{supp}(b_k^{1*})
  \text{ and }
  S_{2k} = \text{supp}(b^2_k) \cup \text{supp}(b_k^{2+}) \cup \text{supp}(b_k^{2*})
\]
capture the support for the $k^{\text{th}}$ column. With this notation, we have
\begin{equation}
\begin{aligned}
\label{eq:one_iteration_basic_inequality}
d^2(B^+, B^*) &= \big\|B_1^+ - B_1^*\big\|_F^2 + \big\|B_2^+ - B_2^*\big\|_F^2 \\
&\leq \xi^2 \Big( \big\| B_1 - B_1^* - \eta \cdot \big[\nabla_{B_1} f\big(B_1, B_2\big) + \nabla_{B_1} g\big(B_1, B_2\big) \big]_{S_1} \big\|_F^2 \\
& \qquad\qquad\qquad + \big\| B_2 - B_2^* - \eta \cdot \big[\nabla_{B_2} f\big(B_1, B_2\big) + \nabla_{B_2} g\big(B_1, B_2\big) \big]_{S_2} \big\|_F^2 \Big) \\
& \leq \xi^2 \Big( d^2(B, B^*) - 2\eta \cdot \big\langle \nabla_{B} f\big(B\big) + \nabla_{B} g\big(B\big), B - B^* \big\rangle_{S_1 \cup S_2} \\
& \qquad\qquad\qquad + \eta^2 \cdot \big\| \big[\nabla_{B} f\big(B\big) + \nabla_{B} g\big(B\big) \big]_{S_1 \cup S_2} \big\|_F^2   \Big) \\
& \leq \xi^2 \Big( d^2(B, B^*) - 2\eta \cdot \big\langle \nabla_{B} f\big(B\big) + \nabla_{B} g\big(B\big), B - B^* \big\rangle_{S_1 \cup S_2} \\
& \qquad\qquad\qquad + 2\eta^2 \cdot \big\| \big[\nabla_{B} f\big(B\big) \big]_{S_1 \cup S_2} \big\|_F^2  +  2\eta^2 \cdot \big\| \big[\nabla_{B} g\big(B\big) \big]_{S_1 \cup S_2} \big\|_F^2   \Big),
\end{aligned}
\end{equation}
where the first inequality follows from Lemma 3.3 of
\cite{li2016stochastic} and
$\xi$ is defined as $\xi^2 = 1 + \frac{2}{\sqrt{c-1}}$ with $c$ set as $s = cs^*$.

Different from the existing work on matrix factorization that focuses
on recovery of a single rank-$K$ matrix, in our model, we have $K$
rank-1 matrices. Therefore we have to deal with each column of $B_1$
and $B_2$ separately. With some abuse of notation, we denote
$f_k(b_k) = f_k(b_k^1, b_k^2) = f_k(\Theta_k) = f(\Theta_1, \ldots, \Theta_k, \ldots, \Theta_K)$
as a function of the $k^{\text{th}}$ columns of $B_1, B_2$, with all the other columns
fixed.
The gradient of $f_k(\Theta_k)$ with respect to $b_k^1$ is then given
by $\nabla f_k(\Theta_k)\cdot b_k^2$. Similarly, we
denote
\begin{equation}
g_k(b_k) = g_k(b_k^1, b_k^2) = \frac{1}{4} \Big( \big\|b^1_k\big\|_2^2 - \big\|b^2_k \big\|_2^2 \Big)^2,
\end{equation}
such that $g(B_1,B_2) = \sum_{k=1}^K g_k(b_k)$.

We first deal the terms involving regularization $g(\cdot)$ in \eqref{eq:one_iteration_basic_inequality}.
Denote $\Delta b_k = \big\|b^1_k\big\|_2^2 - \big\|b^2_k \big\|_2^2$, so that $g_k(b_k) = \frac 14 (\Delta b_k)^2$.
Then
\begin{equation}
\label{eq:upper_bound_G2}
\Big\| \big[ \nabla_{B} g\big(B\big) \big]_{S_1 \cup S_2} \Big\|_F^2  \leq \sum_{k=1}^K \| \nabla g_k(b_k) \|_F^2 \leq
\sum_{k=1}^K (\Delta b_k) ^2 \cdot \| b_k \|_2^2
\leq \|B\|_2^2 \cdot \sum_{k=1}^K (\Delta b_k) ^2.
\end{equation}
Equation (36) in the proof of Lemma B.1 in \cite{park2016finding} gives us
\begin{equation}
\label{eq:g_inner_product}
\big\langle \nabla_{B} g\big(B\big), B - B^* \big\rangle_{S_1 \cup S_2}
\geq \sum_{k=1}^K \Big[ \frac 58 (\Delta b_k) ^2  - \frac 12 \Delta b_k \cdot \| b_k - b_k^* \|_2^2 \Big].
\end{equation}
We then bound the two terms in \eqref{eq:g_inner_product}.
For the first term, we have
\begin{equation}
\begin{aligned}
\label{eq:delta_bk_2}
(\Delta b_k)^2 &\geq
\big\| b_k^1 {b_k^1}^\top - {b_k^1}^* {{b_k^1}^*}^\top \big\|_F^2 + \big\| b_k^2 {b_k^2}^\top - {b_k^2}^* {{b_k^2}^*}^\top \big\|_F^2
- 2 \big\| b_k^1 {b_k^2}^\top - {b_k^1}^* {{b_k^2}^*}^\top \big\|_F^2 \\
&\geq \gamma \cdot \Big( \big\| b_k^1 {b_k^1}^\top - {b_k^1}^* {{b_k^1}^*}^\top \big\|_F^2 + \big\| b_k^2 {b_k^2}^\top - {b_k^2}^* {{b_k^2}^*}^\top \big\|_F^2 + 2 \big\| b_k^1 {b_k^2}^\top - {b_k^1}^* {{b_k^2}^*}^\top \big\|_F^2 \Big) \\
& \qquad\qquad - \frac{4\mu L}{\mu + L} \big\| b_k^1 {b_k^2}^\top - {b_k^1}^* {{b_k^2}^*}^\top \big\|_F^2 \\
& \geq \frac{3}{2} \gamma \big\| \Theta_k^* \big\|_2 \cdot \Big( \big\| b_k^1 - {b_k^1}^* \big\|_2^2 + \big\| b_k^2 - {b_k^2}^* \big\|_2^2 \Big)
 - \frac{4\mu L}{\mu + L} \big\| b_k^1 {b_k^2}^\top - {b_k^1}^* {{b_k^2}^*}^\top \big\|_F^2,
\end{aligned}
\end{equation}
where the last inequality follows from Lemma 5.1 in \cite{tu2016low},
and $\gamma = \min\{1, \mu L/(\mu + L)\}$ as before.
For the second term in \eqref{eq:g_inner_product},
recall that the current iterate satisfies the condition \eqref{eq:bound_current_iterate},
so that
\begin{equation}
\begin{aligned}
\label{eq:delta_bk_diff_bk}
\frac 12 \Delta b_k \cdot \| b_k - b_k^* \|_2^2
 & \leq \frac 12 \Delta b_k \cdot \| b_k - b_k^* \|_2 \cdot \sqrt{ \frac{1}{4} \gamma \sigma^* }\\
&\leq \frac{1}{16} \gamma \sigma^*  \cdot \| b_k - b_k^* \|_2^2 + \frac 14 (\Delta b_k)^2.
\end{aligned}
\end{equation}
Plugging \eqref{eq:delta_bk_diff_bk} and \eqref{eq:delta_bk_2} into
\eqref{eq:g_inner_product} and summing over $k$, we obtain
\begin{equation}
\begin{aligned}
\label{eq:upper_bound_G1}
\big\langle \nabla_{B} g\big(B\big), B - B^* \big\rangle_{S_1 \cup S_2}
&\geq \frac 38 \sum_{k=1}^K (\Delta b_k) ^2  - \frac{1}{16} \sum_{k=1}^K \gamma \sigma^*  \cdot \| b_k - b_k^* \|_2^2 \\
&= \frac 14 \sum_{k=1}^K (\Delta b_k) ^2 + \frac 18 \sum_{k=1}^K (\Delta b_k) ^2 - \frac{1}{16} \gamma \sigma^*  \cdot d^2(B, B^*) \\
&\geq \frac{1}{8} \gamma \sigma^* d^2(B, B^*) - \frac{\mu L}{2(\mu + L)} \big\| b_k^1 {b_k^2}^\top - {b_k^1}^* {{b_k^2}^*}^\top \big\|_F^2 + \frac 14 \sum_{k=1}^K (\Delta b_k) ^2.
\end{aligned}
\end{equation}
Together with \eqref{eq:upper_bound_G2}, we obtain
\begin{equation}
\begin{aligned}
\label{eq:one_iterate_G}
&-2\eta \big\langle \nabla_{B} g\big(B\big), B - B^* \big\rangle_{S_1 \cup S_2} + 2\eta^2 \Big\| \big[ \nabla_{B} g\big(B\big) \big]_{S_1 \cup S_2} \Big\|_F^2 \\
&\qquad \leq -\frac{1}{4} \eta \gamma \sigma^* d^2(B, B^*) + \eta \frac{\mu L}{\mu + L} \big\| b_k^1 {b_k^2}^\top - {b_k^1}^* {{b_k^2}^*}^\top \big\|_F^2 + \Big( 2\eta^2 \|B\|_2^2 - \frac 12 \eta \Big) \sum_{k=1}^K (\Delta b_k) ^2.
\end{aligned}
\end{equation}

Next, we upper bound the terms in
\eqref{eq:one_iteration_basic_inequality} involving the objective
function $f(\cdot)$. For the inner product term, for each $k$, we have
\begin{equation}
\begin{aligned}
&\Big\langle [ \nabla f_k(b_k^1 {b_k^2}^\top) \cdot b_k^2 ]_{S_1}, b_k^1 - {b_k^1}^* \Big\rangle
+ \Big\langle [ \nabla f_k(b_k^1 {b_k^2}^\top) \cdot b_k^1 ]_{S_2}, b_k^2 - {b_k^2}^* \Big\rangle \\
&= \Big\langle \nabla f_k(b_k^1 {b_k^2}^\top), (b_k^1 - {b_k^1}^*){b_k^2}^\top + b_k^1 ({b_k^2 - {b_k^2}^*})^\top \Big\rangle_{S_{1k}, S_{2k}} \\
&= \Big\langle \nabla f_k(b_k^1 {b_k^2}^\top), (b_k^1 - {b_k^1}^*)({b_k^2 - {b_k^2}^*})^\top + b_k^1{b_k^2}^\top - {b_k^1}^* {{b_k^2}^*}^\top \Big\rangle_{S_{1k}, S_{2k}} \\
&= \Big\langle \nabla f_k(b_k^1 {b_k^2}^\top), (b_k^1 - {b_k^1}^*)({b_k^2 - {b_k^2}^*})^\top \Big\rangle_{S_{1k}, S_{2k}}
 + \Big\langle \nabla f_k(b_k^1 {b_k^2}^\top), b_k^1{b_k^2}^\top - {b_k^1}^* {{b_k^2}^*}^\top \Big\rangle_{S_{1k}, S_{2k}} \\
&= \underbrace{\Big\langle \nabla f_k(b_k^1 {b_k^2}^\top), (b_k^1 - {b_k^1}^*)({b_k^2 - {b_k^2}^*})^\top \Big\rangle_{S_{1k}, S_{2k}} }_{W_{1k}}
 + \underbrace{ \Big\langle \nabla f_k( {b_k^1}^* {{b_k^2}^*}^\top), b_k^1{b_k^2}^\top - {b_k^1}^* {{b_k^2}^*}^\top \Big\rangle_{S_{1k}, S_{2k}} }_{W_{2k}} \\
& \qquad\qquad\qquad + \underbrace{ \Big\langle \nabla f_k(b_k^1 {b_k^2}^\top) - \nabla f_k( {b_k^1}^* {{b_k^2}^*}^\top), b_k^1{b_k^2}^\top - {b_k^1}^* {{b_k^2}^*}^\top \Big\rangle_{S_{1k}, S_{2k}} }_{W_{3k}}.
\end{aligned}
\end{equation}
For the term $W_{3k}$, Theorem 2.1.11 of \cite{Nesterov2013Introductory} gives
\begin{equation}
\begin{aligned}
\label{eq:bound_W3k}
W_{3k} & \geq
\frac{\mu L}{\mu + L }\cdot\Big\|{b_k^1 {b_k^2}^\top - {b_k^1}^* {{b_k^2}^*}^\top }\Big\|_F^2
 + \frac{1}{\mu +L }\cdot\Big\|\sbr{\nabla f( b_k^1 {b_k^2}^\top ) - \nabla f( {b_k^1}^* {{b_k^2}^*}^\top )}_{S_{1k},S_{2k}}\Big\|_F^2.
\end{aligned}
\end{equation}
For the term $W_{2k}$, according to the definition of the statistical error in \eqref{eq:stat_error_definition}, we have
\begin{equation}
\begin{aligned}
\label{eq:bound_W2k}
\sum_{k=1}^K W_{2k} &\geq
-e_{\rm stat}\cdot \sum_{k=1}^K \Big\| b_k^1{b_k^2}^\top - {b_k^1}^* {{b_k^2}^*}^\top \Big\|_F \\
&\geq - \frac{K}{2}\frac{\mu + L }{\mu L}e_{\rm stat}^2 -
\frac{1}{2}\frac{\mu L}{\mu + L } \sum_{k=1}^K \Big\|{ b_k^1{b_k^2}^\top - {b_k^1}^* {{b_k^2}^*}^\top }\Big\|_F^2.
\end{aligned}
\end{equation}
For the term $W_{1k}$,
\begin{equation}
\begin{aligned}
\label{eq:bound_W1k}
\sum_{k=1}^K W_{1k} &= \sum_{k=1}^K
\Big\langle \nabla f_k( {b_k^1}^* {{b_k^2}^*}^\top ), (b_k^1 - {b_k^1}^*)({b_k^2 - {b_k^2}^*})^\top \Big\rangle_{S_{1k}, S_{2k}}\\
&\qquad\qquad\qquad + \Big\langle \nabla f_k(b_k^1 {b_k^2}^\top) - \nabla f_k( {b_k^1}^* {{b_k^2}^*}^\top), (b_k^1 - {b_k^1}^*)({b_k^2 - {b_k^2}^*})^\top \Big\rangle_{S_{1k}, S_{2k}} \\
& \geq - \rbr{e_{\rm stat} + \sum_{k=1}^K \Big\|\sbr{\nabla f_k(b_k^1 {b_k^2}^\top) - \nabla f_k( {b_k^1}^* {{b_k^2}^*}^\top)}_{S_{1k},S_{2k}}\Big\|_F}\cdot d^2(B, B^*) \\
& \geq - \rbr{e_{\rm stat} + \sum_{k=1}^K\Big\|\sbr{\nabla f_k(b_k^1 {b_k^2}^\top) - \nabla f_k( {b_k^1}^* {{b_k^2}^*}^\top)}_{S_{1k},S_{2k}}\Big\|_F}
\sqrt{\frac{\gamma \sigma^*}{16(\mu +L )}} d(B, B^*) \\
& \geq -\frac{K}{2(\mu +L )}\cdot\rbr{e_{\rm stat}^2
 + \sum_{k=1}^K \Big\|\sbr{\nabla f_k(b_k^1 {b_k^2}^\top) - \nabla f_k( {b_k^1}^* {{b_k^2}^*}^\top)}_{S_{1k},S_{2k}}\Big\|_F^2} \\
&\qquad\qquad\qquad\qquad\qquad\qquad\qquad\qquad\qquad\qquad\qquad\qquad
- \frac{1}{16} \gamma \sigma^*
\cdot d^2(B, B^*),
\end{aligned}
\end{equation}
where we use the fact that $d(B,B^*)$ satisfies \eqref{eq:bound_current_iterate},
\begin{equation}
\big\|(b_k^1 - {b_k^1}^*)({b_k^2 - {b_k^2}^*})^\top\big\|_F \leq \big\|b_k^1 - {b_k^1}^*\big\|_F \big\|{b_k^2 - {b_k^2}^*}\big\|_F \leq  \big\|b_k^1 - {b_k^1}^*\big\|_F^2 + \big \|{b_k^2 - {b_k^2}^*} \big\|_F^2,
\end{equation}
and that their summation is $d^2(B, B^*)$.
For the term in \eqref{eq:one_iteration_basic_inequality}
involving square of $f(\cdot)$, we have
\begin{equation}
\label{eq:bound_F2}
\Big\| \big[\nabla_{B} f\big(B\big) \big]_{S_1 \cup S_2} \Big\|_F^2  \leq
4\cdot
\Big( \sum_{k=1}^K
\Big\|\sbr{\nabla f(b_k^1 {b_k^2}^\top)-\nabla f({b_k^1}^* {{b_k^2}^*}^\top )}_{S_1, S_2}\Big\|_F^2 +
e_{\text{stat}}^2
\Big) \cdot \|B\|_{2}^2.
\end{equation}
Combining \eqref{eq:bound_W3k},  \eqref{eq:bound_W2k}, \eqref{eq:bound_W1k}, and  \eqref{eq:bound_F2}, we obtain
\begin{equation}
\begin{aligned}
\label{eq:one_iterate_F}
&-2\eta \big\langle \nabla_{B} f\big(B\big), B - B^* \big\rangle_{S_1 \cup S_2} + \eta^2 \Big\| \big[\nabla_{B} f\big(B\big) \big]_{S_1 \cup S_2} \Big\|_F^2 \\
&\leq
e_{\text{stat}}^2 \cdot \Big( 8\|B\|_2^2\eta^2 + \frac{K(\mu+L)}{\mu L} \eta + \frac{K}{\mu+L} \eta \Big) \\
& \qquad\qquad- \frac{\mu L }{\mu + L } \eta \sum_{k=1}^K \Big\|{ b_k^1{b_k^2}^\top - {b_k^1}^* {{b_k^2}^*}^\top }\Big\|_F^2 + \frac{1}{8} \gamma \sigma^*\eta \cdot d^2(B, B^*) \\
&\qquad\qquad + \Big(8\eta^2 \|B\|_2^2 - \frac{K\eta}{\mu + L} \Big) \sum_{k=1}^K \Big\|\sbr{\nabla f_k(b_k^1 {b_k^2}^\top) - \nabla f_k( {b_k^1}^* {{b_k^2}^*}^\top)}_{S_{1k},S_{2k}}\Big\|_F^2.
\end{aligned}
\end{equation}
Plugging \eqref{eq:one_iterate_G} and \eqref{eq:one_iterate_F} into \eqref{eq:one_iteration_basic_inequality}, we obtain
\begin{equation}
\begin{aligned}
\label{eq:one_iterate_inequality_full}
d^2(B^+, B^*) &=
\xi^2 \rbr{1 - \frac{1}{4}  \gamma \sigma^*\eta} \cdot d^2(B, B^*) + \xi^2 \Big( 2\eta^2 \|B\|_2^2 - \frac 12 \eta \Big) \sum_{k=1}^K (\Delta b_k) ^2 \\
& \qquad +
\xi^2 \rbr{8\eta^2 \cdot\|B\|_2^2 - \frac{K\eta}{\mu + L}}\cdot \sum_{k=1}^K \Big\|\sbr{\nabla f_k(b_k^1 {b_k^2}^\top) - \nabla f_k( {b_k^1}^* {{b_k^2}^*}^\top)}_{S_{1k},S_{2k}}\Big\|_F^2 \\
&\qquad +
\xi^2 \rbr{
\frac{K(\mu+L)}{\mu L} \eta +\frac{K\eta}{\mu + L} + 8 \eta^2 \cdot\|B\|_2^2
}\cdot e_{\rm stat}^2.
\end{aligned}
\end{equation}
When the step size satisfies
\begin{equation}
\label{eq:choice_of_step_size}
\eta \leq \frac{1}{4\|B\|_2^2} \cdot \min\Big\{\frac{K}{2(\mu +L )}, 1\Big\},
\end{equation}
the second and third terms in \eqref{eq:one_iterate_inequality_full} are non-positive.
Therefore, we can upper bound them with $0$ to obtain
\begin{equation}
\label{eq:bound_one_iterate}
d^2 \Big( B^{(t+1)} , B^* \Big) \leq \beta\cdot d^2 \Big( B^{(t)}, B^* \Big) + 3\eta K \xi^2 \cdot\frac{\mu + L }{\mu L } \cdot e_{{\rm stat}}^2 ,
\end{equation}
with the contraction value
\begin{equation}
\label{eq:def_beta_appendix}
\beta = \xi^2 \Big( 1 - \frac{1}{4}  \gamma \sigma^*\eta \Big) < 1.
\end{equation}

From \eqref{eq:def_beta_appendix} we see that $\beta$ is a
multiplication of two terms. The first term
$\xi^2 = 1 + \frac{2}{\sqrt{c-1}}$ is slightly larger than 1, while
the second term is smaller than 1.  In order to guarantee that
$\beta < 1$, we should choose a conservative hard thresholding
parameter (recall that $s = c \cdot s^*$), so that $\xi^2$ is close to
1. In practice, we observe that $\beta < 1$ for a large range of hard
thresholding parameters.  Notice that without the hard thresholding
step, we are guaranteed to have $\beta < 1$.

In order to iteratively apply the error bound
\eqref{eq:bound_one_iterate}, we need to show that the condition
\eqref{eq:bound_current_iterate} is satisfied in each iteration.  A
sufficient condition is to require
\begin{equation}
\label{eq:ass_estat}
e_{\rm stat}^2 \leq \frac{1-\beta}{3\eta K \xi^2} \cdot \frac{\mu L }{\mu + L }\cdot  { \frac{1}{4} \gamma \sigma^* } \cdot \min \Big\{ 1, \frac{1}{4(\mu +L )} \Big\}.
\end{equation}
It is straightforward to verify that \eqref{eq:bound_one_iterate} and
\eqref{eq:ass_estat} imply that the next iterate also satisfies the
condition \eqref{eq:bound_one_iterate}.  To justify the condition
\eqref{eq:ass_estat}, consider the case where the condition
\eqref{eq:ass_estat} is violated. Together with
\eqref{eq:bound_one_iterate}, this shows that
$d^2(B,B^*) \leq C \cdot e_{\rm stat}^2$, which means that the current
iterate is already optimal. Therefore, we can assume
\eqref{eq:ass_estat} and then \eqref{eq:bound_one_iterate} is
satisfied for all the iterations.

With the error bound \eqref{eq:bound_one_iterate} we can complete the
proof. For a large enough sample size, the initial point $B^{(0)}$
satisfies \eqref{eq:bound_current_iterate}. The proof above shows that
\eqref{eq:bound_one_iterate} is satisfied with $t = 0$.  The condition
\eqref{eq:ass_estat} ensures that the next iterate $B^{(1)}$ also
satisfies \eqref{eq:bound_current_iterate}.  Iterating the argument,
we obtain
\begin{equation}
d^2 \Big( B^{(T)}, B^* \Big) \leq \beta^T \cdot d^2 \Big( B^{(0)}, B^* \Big) +  \frac{3\eta K\xi^2}{1-\beta}\cdot\frac{\mu + L }{\mu L}\cdot e_{{\rm stat}}^2,
\end{equation}
which shows that the iterates of Algorithm \ref{algo_2} converge
linearly to the true value up to a statistical error.

Finally, it remains to provide an upper bound on the step size
\eqref{eq:choice_of_step_size} that is independent of the norm of
the value in each iterate $\|B\|_2$, as given in
\eqref{eq:step_size_selection_fixed}. This can be established as
in the proof of Lemma 4 in \cite{yu2020recovery}.
The proof is now complete.

\newpage
\bibliography{19-496}

\end{document}